\documentclass{article}

\PassOptionsToPackage{numbers, compress}{natbib}

\usepackage[final]{neurips_2020}

\usepackage[utf8]{inputenc} 
\usepackage{lipsum} 

\usepackage[T1]{fontenc}

\usepackage{etoolbox}
\newbool{includeappendix}
\setbool{includeappendix}{true}

\ifdefined\isoverfull
\else
\fi

\newcommand{\eg}{e.g., }
\newcommand{\ie}{i.e., }

\newcommand{\cf}{{cf.\ }}

\usepackage{acro} 

\DeclareAcronym{cli} {
    short = CLI,
    long = Command Line Interface,
    class = abbrev
}

\usepackage{listings}

\usepackage{textcomp}

\usepackage{xcolor}

\usepackage[scaled=0.8]{beramono}

\definecolor{ckeyword}{HTML}{7F0055}
\definecolor{ccomment}{HTML}{3F7F5F}
\definecolor{cstring}{HTML}{2A0099}

\lstdefinestyle{numbers}{
	numbers=left,
	framexleftmargin=20pt,
	numberstyle=\tiny,
	firstnumber=auto,
	numbersep=1em,
	xleftmargin=2em
}

\lstdefinestyle{layout}{
	frame=none,
	captionpos=b,
}

\lstdefinestyle{comment-style}{
	morecomment=[l]//,
	morecomment=[s]{/*}{*/},
	commentstyle={\color{ccomment}\itshape},
}

\lstdefinestyle{string-style}{
	morestring=[b]",%
	morestring=[b]',%
	stringstyle={\color{cstring}},
	showstringspaces=false,%
}

\lstdefinestyle{keyword-style}{
	keywordstyle={\ttfamily\bfseries},
	morekeywords={
		function,
		constructor,
		int,
		bool,
		return,
		returns,
		uint
	},
	morekeywords = [2]{},
	keywordstyle = [2]{\text},
	sensitive=true,
}

\lstdefinestyle{input-encoding}{
	inputencoding=utf8,
	extendedchars=true,
	literate=
	{ℝ}{$\reals$}1%
	{→}{$\rightarrow$}1%
	{α}{$\alpha$}1%
	{β}{$\beta$}1%
	{λ}{$\lambda$}1%
	{θ}{$\theta$}1%
	{ϕ}{$\phi$}1%
}

\lstdefinestyle{escaping}{
	moredelim={**[is][\color{blue}]{\%}{\%}},
	escapechar=|,
	mathescape=true
}

\lstdefinestyle{default-style}{
	basicstyle=\fontencoding{T1}\ttfamily\footnotesize,
	style=numbers,
	style=layout,
	style=comment-style,
	style=string-style,
	style=keyword-style,
	style=input-encoding,
	style=escaping,
	tabsize=2,
	upquote=true
}

\lstdefinelanguage{BASIC}{
	language=C++,
	style=default-style
}[keywords,comments,strings]%

\usepackage{tikz}

\usetikzlibrary{arrows}
\usetikzlibrary{automata}
\usetikzlibrary{calc}
\usetikzlibrary{backgrounds}
\usetikzlibrary{decorations.markings}
\usetikzlibrary{decorations.pathmorphing}
\usetikzlibrary{decorations.pathreplacing}
\usetikzlibrary{fit}
\usetikzlibrary{patterns}
\usetikzlibrary{positioning}
\usetikzlibrary{shadows}
\usetikzlibrary{shapes}
\usetikzlibrary{shapes.geometric}

\usepackage{microtype}
\usepackage{graphicx}
\usepackage{subfigure}
\usepackage{booktabs} 
\usepackage{adjustbox}
\usepackage{amssymb}
\usepackage{amsmath}
\usepackage{amsthm}
\usepackage{amsfonts}
\usepackage{url}
\usepackage{hyperref}
\usepackage{nicefrac}

\usepackage[capitalize, noabbrev]{cleveref}

\newcommand{\app}[1]{%
\ifbool{includeappendix}{\cref{#1}}{the appendix}%
}
\newcommand{\App}[1]{%
\ifbool{includeappendix}{\cref{#1}}{The appendix}%
}

\usepackage{pgfplots}

\usepackage{bbm}
\usepackage{multirow}

\DeclareMathOperator*{\argmin}{arg\,min}
\newtheorem{theorem}{Theorem}

\title{Learning Certified Individually Fair Representations}

\author{
    Anian Ruoss, Mislav Balunovi\'{c}, Marc Fischer, Martin Vechev \\
    Department of Computer Science \\
    ETH Zurich \\
    \texttt{anruoss@ethz.ch} \\
    \texttt{\{mislav.balunovic, marc.fischer, martin.vechev\}@inf.ethz.ch}
}

\begin{document}

    \maketitle


    \begin{abstract}
        Fair representation learning provides an effective way of enforcing fairness
constraints without compromising utility for downstream users.
A desirable family of such fairness constraints, each requiring similar
treatment for similar individuals, is known as individual fairness.
In this work, we introduce the first method that enables data consumers to
obtain certificates of individual fairness for existing and new data points.
The key idea is to map similar individuals to close latent representations and
leverage this latent proximity to certify individual fairness.
That is, our method enables the data producer to learn and certify a
representation where for a data point all similar individuals are at
$\ell_\infty$-distance at most $\epsilon$, thus allowing data consumers to
certify individual fairness by proving $\epsilon$-robustness of their
classifier.
Our experimental evaluation on five real-world datasets and several fairness
constraints demonstrates the expressivity and scalability of our approach.

    \end{abstract}

    \section{Introduction}
\label{sec:introduction}

The increased use of machine learning in sensitive domains (e.g., crime risk
assessment~\cite{brennan2009compas}, ad targeting~\cite{datta2015automated},
and credit scoring~\cite{khandani2010consumer}) has raised concerns that
methods learning from data can reinforce human bias, discriminate, and
lack fairness~\cite{sweeney2013discrimination,bolukbasi2016man,barocas2016big}.
Moreover, data owners often face the challenge that their data will
be used in (unknown) downstream applications, potentially indifferent to
fairness concerns~\cite{cisse2019fairness}.
To address this challenge, the paradigm of learning fair representations has
emerged as a promising approach to obtain data representations that preserve
fairness while maintaining utility for a variety of downstream
tasks~\cite{zemel2013learning, madras2018learning}.
The recent work of~\citet{mcnamara2017provably} has formalized this setting by
partitioning the landscape into: a \emph{data regulator} who
defines fairness for the particular task at hand, a \emph{data producer} who
processes sensitive user data and transforms it into another representation,
and a \emph{data consumer} who performs predictions based on the new
representation.

In this setting, a machine learning model
$M \colon \mathbb{R}^n \rightarrow \mathbb{R}^o$ is composed of two parts: an
encoder $f_\theta \colon \mathbb{R}^n \rightarrow \mathbb{R}^k$, provided by
the data producer, and a classifier
$h_\psi \colon \mathbb{R}^k \rightarrow \mathbb{R}^o$, provided by the data
consumer, with $\mathbb{R}^k$ denoting the latent space.
The data regulator selects a definition of fairness that the model $M$ should
satisfy.
Most work so far has explored two main families of fairness
definitions~\cite{chouldechova2018frontiers}: \emph{statistical} and
\emph{individual}.
Statistical notions define specific groups in the population and require that
particular statistics, computed based on model decisions, should be equal for
all groups.
Popular notions of this kind include demographic
parity~\cite{dwork2012fairness} and equalized odds~\cite{hardt2016equality}.
While these notions do not require any assumptions on the data and are easy to
certify, they offer no guarantees for individuals or other subgroups in the
population~\cite{kearns2018preventing}.
In contrast, individual notions of fairness~\cite{dwork2012fairness} are
desirable as they explicitly require that similar individuals in the population
are treated similarly.

\paragraph{Key challenge}

A central challenge then is to enforce individual fairness in the setting
described above.
That is, to both learn an individually fair representation and to certify that
individual fairness is actually satisfied across the end-to-end model $M$
without compromising the independence of the data producer and the data
consumer.

\paragraph{This work}

In this work, we propose the first method for addressing the above challenge.
At a high level, our approach is based on the observation that recent advances
in training machine learning models with logical
constraints~\cite{fischer2019dl2} together with new methods for proving that
constraints are satisfied~\cite{tjeng2017evaluating} open the possibility for
learning certified individually fair models.

Concretely, we identify a practical class of individual fairness
definitions captured via declarative fairness constraints.
Such a fairness constraint is a binary similarity function
$\phi \colon \mathbb{R}^n \times \mathbb{R}^n \rightarrow \{ 0, 1 \}$,
where $\phi(x, x')$ evaluates to 1 if and only if two individuals $x$
and $x'$ are similar (\eg if all their attributes except for race are
the same).
By working with declarative constraints, data regulators can now express
interpretable, domain-specific notions of similarity, a problem known to be
challenging~\cite{zemel2013learning, lahoti2019ifair,
lahoti2019operationalizing, yurochkin2020training, wang2019empirical,
ilvento2020metric}.

Given the fairness constraint $\phi$, we can now train an individually fair
representation and use it to obtain a certificate of individual fairness for
the end-to-end model.
For training, the data producer can employ our framework to learn an encoder
$f_\theta$ with the goal that two individuals satisfying $\phi$ should be
mapped close together in $\ell_{\infty}$-distance in latent space.
As a consequence, individual fairness can then be certified for a data point in
two steps: first, the data producer computes a convex relaxation of the latent
set of similar individuals and passes it to the data consumer.
Second, the data consumer certifies individual fairness by proving local
robustness within the convex relaxation.
Importantly, the data consumer can now perform \emph{modular} certification: it
does not need to know the fairness constraint $\phi$ and the concrete data
point $x$.

Our experimental evaluation on several datasets and fairness constraints shows
a substantial increase (up to 72.6\%) of certified individuals (unseen during
training) when compared to standard representation learning.

\paragraph{Main contributions}

Our key contributions are:
\begin{itemize}
    \item A practical family of similarity notions for individual fairness
        defined via interpretable logical constraints.
    \item A method to learn individually fair representations (defined in an
        expressive logical fragment), which comes with provable certificates.
    \item An end-to-end implementation of our method in an open-source tool
        called LCIFR, together with an extensive evaluation on several
        datasets, constraints, and architectures.
        We make LCIFR publicly available at \url{https://github.com/eth-sri/lcifr}.
\end{itemize}

    \section{Overview}
\label{sec:overview}

\begin{figure*}
    \centering
    \begin{adjustbox}{width=\textwidth}
        \begin{tikzpicture}

    \definecolor{netborder}{RGB}{31,120,180}
    \definecolor{netinside}{RGB}{166,206,227}

    \definecolor{Sborder}{RGB}{  150, 109, 80}
    \definecolor{Sinside}{RGB}{184, 138, 106}

    \definecolor{Pinside}{RGB}{230, 150, 71}
    \definecolor{Pborder}{RGB}{186, 109, 34}

    \definecolor{DPinside}{RGB}{151, 155, 156}

    \definecolor{Correct}{RGB}{140, 191, 94}
    \definecolor{Wrong}{RGB}{200, 43, 17}


    \node[] (dp) at (3.7, 2.4) {Data producer};
    \draw[thick, opacity=0.2, fill=DPinside, draw=DPinside] (-0.2, -0.2) -- (-0.2, 2.2) -- (7.25, 2.2) -- (7.25, -0.2) -- (-0.2, -0.2);

    \node[] (x1) at (1, 1.3) {};

    \draw[->, thick] (0.4, 0.25) -- (1.65, 0.25);
    \node[] (axisx1) at (1.65, 0.0) {$x_1$};
    \draw[->, thick] (0.4, 0.25) -- (0.4, 1.85);
    \node[] (axisx2) at (0.1, 1.85) {$x_2$};
    \draw[thick, fill=Sinside, draw=Sborder] (0.5, 0.3) -- ++(0, 1) -- ++(0.5, 0.5) -- ++(0.5, -1) -- ++(-0.5, 0.2) -- ++(-0.5, -0.7);
    \filldraw[black] (x1) circle (1.5pt);

    \node[font=\footnotesize] (sphi) at (0.9, -0.8) {$S_\phi \left( x \right)$};
    \node[font=\footnotesize] (x) at (0.8, 1.2) {$x$};

    \node[rectangle, draw, minimum width=2cm, align=center, very thick, color=netborder, fill=netinside, text=black, rotate=90]
    at (2.25, 1) {ReLU};
    \draw[->] (1.6, 1) -- (1.9, 1);

    \node[rectangle, draw, minimum width=2cm, align=center, very thick, color=netborder, fill=netinside, text=black, rotate=90]
    at (3.25, 1) {ReLU};
    \draw[->] (2.6, 1) -- (2.9, 1);

    \node[rectangle, draw, minimum width=1cm, align=center, very thick, color=netborder, fill=netinside, text=black, rotate=90]
    at (4.25, 1) {FC};
    \draw[->] (3.6, 1) -- (3.9, 1);

    \draw[thick, decoration={brace, mirror, raise=10pt}, decorate]
    (1.9,0) -- node[below=15pt, font=\footnotesize] {$f_\theta$} (4.6,0);

    \node[align=center, font=\footnotesize] (latent) at (6.1, -0.75) {
    $f_\theta \left( S_\phi \left( x \right) \right)$\\
    $\subseteq \mathbb{B}_\infty \left( z, \epsilon \right)$};

    \node[] (fx1) at (6.25, 1.0) {};
    \draw[->] (4.6, 1) -- (5.2, 1);
    \draw[->, thick] ($(fx1) + (-0.95, -0.75)$) -- ++(0, 1.5);
    \node[] (axisz1) at ($(fx1) + (-0.95, -0.75) + (-0.3, 1.5)$) {$z_2$};
    \draw[->, thick] ($(fx1) + (-0.95, -0.75)$) -- ++(1.5, 0);
    \node[] (axisz1) at ($(fx1) + (-0.95, -0.75) + (1.5, -0.25)$) {$z_1$};

    \draw[very thick, dashed, fill=Pinside, draw=Pborder] ($(fx1) + (-0.5, -0.5)$) -- ++(0, 2*0.5) -- ++(2*0.5, 0) -- ++(0, -2*0.5) -- ++(-2*0.5, 0);

    \draw[thick, fill=Sinside, draw=Sborder] (5.8, 0.5) -- ++(0.1, 0.5) -- ++(0.5, 0.4) -- ++(0.1, -0.4) -- ++(0.2, -0.2) -- ++(-0.9, -0.3);
    \filldraw[black] (fx1) circle (1.5pt);

    \node[font=\footnotesize] (z) at (6.05, 0.9) {$z$};


    \node[] (dp) at (11.1, 2.4) {Data consumer};
    \draw[thick, opacity=0.2, fill=DPinside, draw=DPinside] (7.45, -0.2) -- (7.45, 2.2) -- (14.7, 2.2) -- (14.7, -0.2) -- (7.45, -0.2);

    \node[] (arrowtextabove) at (7.35, 1.25) {$z, \epsilon$};
    \draw[->] (6.9, 1) -- (7.9, 1);

    \node[] (fx2) at (9, 1.0) {};
    \draw[->, thick] ($(fx2) + (-0.95, -0.75)$) -- ++(0, 1.5);
    \node[] (axisz1) at ($(fx2) + (-0.95, -0.75) + (-0.3, 1.5)$) {$z_2$};
    \draw[->, thick] ($(fx2) + (-0.95, -0.75)$) -- ++(1.5, 0);
    \node[] (axisz1) at ($(fx2) + (-0.95, -0.75) + (1.5, -0.25)$) {$z_1$};

    \draw[very thick, dashed, fill=Pinside, draw=Pborder] ($(fx2) + (-0.5, -0.5)$) -- ++(0, 2*0.5) -- ++(2*0.5, 0) -- ++(0, -2*0.5) -- ++(-2*0.5, 0);

    \node[align=center, font=\footnotesize] (latent2) at (8.8, -0.8) {$\mathbb{B}_\infty \left(z, \epsilon \right)$};

    \filldraw[black] (fx2) circle (1.5pt);
    \node[font=\footnotesize] (z) at (8.8, 0.9) {$z$};

    \node[rectangle, draw, minimum width=2cm, align=center, very thick, color=netborder, fill=netinside, text=black, rotate=90]
    at (10.4, 1) {ReLU};
    \draw[->] (9.75, 1) -- (10.05, 1);

    \node[rectangle, draw, minimum width=2cm, align=center, very thick, color=netborder, fill=netinside, text=black, rotate=90]
    at (11.5, 1) {Sigmoid};
    \draw[->] (10.75, 1) -- (11.05, 1);

    \draw[thick, decoration={brace, mirror, raise=10pt}, decorate]
    (10,0) -- node[below=15pt] {$h_\psi$} (12,0);

    \draw[fill=Correct, opacity=0.8, draw=none] (12.5, 0) -- ++(0, 2) -- ++(2, 0) -- ++(0, -1) -- ++(-1, -0.8) -- ++(-1, -0.2);
    \draw[fill=Wrong, opacity=0.8, draw=none] (14.5, 0) -- (14.5, 1.0) -- (13.5, 0.2) -- (12.5, 0.0) -- (14.5, 0.0);

    \node[align=center, font=\footnotesize] (output) at (13.5, -0.8) {$h_\psi \left( \mathbb{B}_\infty \left( z, \epsilon \right) \right)$};

    \draw[thick, fill=Pinside, draw=Pborder] (12.6, 0.2) -- ++(0.6, 1.2) -- ++(0.6, 0.4) -- ++(0.15, -0.8) -- ++(0.15, -0.2) -- ++(-1.5, -0.6);
    \node[] (hz) at (13.5, 0.7) {};
    \filldraw[black] (hz) circle (1.5pt);

    \node[font=\footnotesize] (hz) at (13.5, 0.95) {$h_\psi \left( z \right)$};

    \draw[->] (11.85, 1) -- (12.4, 1);

\end{tikzpicture}
    \end{adjustbox}
    \caption{
        Overview of our framework.
        The left side shows the component corresponding to the data producer
        who learns an encoder $f_\theta$ which maps the entire set of
        individuals $S_\phi(x)$ that are similar to individual $x$, according
        to the similarity notion $\phi$, to points near $f_\theta(x)$ in the
        latent space.
        The data producer then computes an $\ell_\infty$-bounding box
        $\mathbb{B}_\infty$ around the latent set of similar individuals
        $f_\theta(S_\phi(x))$ with center $z = f_\theta(x)$ and radius
        $\epsilon$ and passes it to the data consumer.
        The data consumer receives the latent representation $z$ and radius
        $\epsilon$, trains a classifier $h_\psi$, and certifies that the entire
        $\ell_\infty$-ball centered around $z$ with radius $\epsilon$ is
        classified the same (green color shows fair output region).
    }
    \label{fig:overview}
\end{figure*}
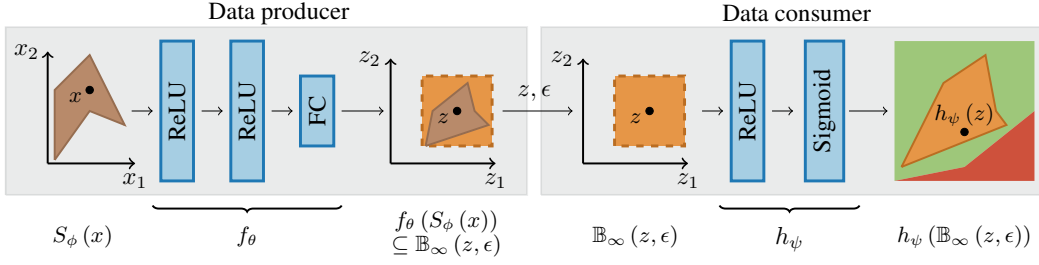

This section provides a high-level overview of our approach, with the
overall flow shown in~\cref{fig:overview}.

As introduced earlier, our setting consists of three parties.
The first party is a data regulator who defines similarity measures for the
input and the output denoted as $\phi$ and $\mu$, respectively.
The properties $\phi$ and $\mu$ are problem-specific and can be expressed in a
rich logical fragment which we describe later in~\cref{sec:training}.
For example, for classification tasks $\mu$ could denote equal classification
(\ie $\mu(M(x), M(x')) = 1 \iff M(x) = M(x')$) or classifying $M(x)$ and
$M(x')$ to the same label group; for regressions tasks $\mu$ could evaluate
to 1 if $\|M(x) - M(x')\| \leq 0.1$ and 0 otherwise.
We focus on equal classification in the classification setting for the
remainder of this work.

The goal of treating similar individuals as similarly as possible can then be
formulated as finding a classifier $M$ which maximizes
\begin{equation}
    \label{eq:fair-risk}
    \quad \mathbb{E}_{x \sim D} \left[
        \forall x' \in \mathbb{R}^n : \phi(x, x') \implies \mu(M(x), M(x'))
    \right],
\end{equation}
where $D$ is the underlying data distribution (we assume a logical expression
evaluates to 1 if it is true and to 0 otherwise).
As usual in machine learning, we approximate this quantity with the empirical
risk, by computing the percentage of individuals $x$ from the test set for
which we can certify that
\begin{equation}
    \label{eq:individual-fairness}
    \forall x' \in S_\phi(x): \mu(M(x), M(x')),
\end{equation}
where $S_\phi(x) = \{x' \in \mathbb{R}^n \mid \phi(x, x')\}$ denotes the set of
all points similar to $x$.
Note that $S_\phi(x)$ generally contains an infinite number of individuals.
In~\cref{fig:overview}, $S_\phi(x)$ is represented as a brown shape, and $x$ is
shown as a single point inside of $S_\phi(x)$.

The key idea of our approach is to train the encoder $f_\theta$ to map point
$x$ and all points $x' \in S_\phi(x)$ close to one another in the latent space
with respect to $\ell_{\infty}$-distance, specified as
\begin{equation}
    \label{eq:encoder_constraint}
    \phi \left( x, x' \right) \implies ||f_\theta(x') - f_\theta(x)||_\infty \leq \delta,
\end{equation}
where $\delta$ is a tunable parameter of the method, determined in agreement
between producer and consumer (we could also use another $\ell_p$-norm).
If the encoder indeed satisfies~\cref{eq:encoder_constraint}, the data
consumer, potentially indifferent to the fairness constraint, can then train a
classifier $h_\psi$ independently of the similarity notion $\phi$.
The data consumer only has to train $h_\psi$ to be robust to
perturbations up to $\delta$ in $\ell_\infty$-norm, which can be solved via
standard min-max optimization, discussed in~\cref{sec:training}.

We now explain our end-to-end inference with provable certificates for encoder
$f_\theta$ and classifier $h_\psi$.

\paragraph{Processing the producer model}

Given a data point $x$, we first propagate both $x$ and its set of
similar points $S_\phi(x)$ through the encoder, as shown
in~\cref{fig:overview}, to obtain the latent representations
$z = f_\theta(x)$ and $f_\theta(S_\phi(x))$.  As
\cref{eq:encoder_constraint} may not hold for the particular $x$ and
$\delta$ due to the stochastic nature of training, we compute the
smallest $\ell_\infty$-bounding box of radius $\epsilon$ such that
$f_\theta(S_\phi(x)) \subseteq \mathbb{B}_\infty(z, \epsilon) := \{ z'
\mid \|z - z'\|_{\infty} \leq \epsilon \}$.  This
$\ell_\infty$-bounding box with center $z$ and radius $\epsilon$ is
shown as orange in~\cref{fig:overview}.

\paragraph{Processing the consumer model}

Next, we provide the latent representation $z$ and the radius $\epsilon$ to the
data consumer.
The data consumer then knows that all points similar to $x$ are in the
$\ell_\infty$-ball of radius $\epsilon$, but does not need to know the
similarity constraint $\phi$ nor the particular shape $f_\theta(S_\phi(x))$.
The key observation is the following: if the data consumer can prove its
classifier $h_\psi$ is robust to $\ell_\infty$-perturbations up to $\epsilon$
around $z$, then the end-to-end classifier $M = h_\psi \circ f_\theta$
satisfies individual fairness at $x$ with respect to the similarity rule $\phi$
imposed by the data regulator.

There are two central technical challenges we need to address.
The first challenge is how to train an encoder to
satisfy~\cref{eq:encoder_constraint}, while not making any domain-specific
assumptions about the point $x$ or the similarity constraint $\phi$.
The second challenge is how to provide a certificate of individual fairness for
$x$, which requires both computing the smallest radius $\epsilon$ such that
$f_\theta(S_\phi(x)) \subseteq \mathbb{B}_\infty(z, \epsilon)$, as well as
certifying $\ell_\infty$-robustness of the classifier $h_\psi$.

To train an encoder, we build on~\citet{fischer2019dl2}, who provide a
translation from logical constraints $\phi$ to a differentiable loss function.
The training of the encoder network can then be formulated as a min-max
optimization problem, which alternates between (i) searching for
counterexamples $x' \in S_\phi(x)$ that violate~\cref{eq:encoder_constraint},
and (ii) training $f_\theta$ on the counterexamples.
We employ gradient descent to minimize a joint objective composed of a
classification loss and the constraint loss obtained from
translating~\cref{eq:encoder_constraint}.
Once no more counterexamples are found, we can conclude the encoder empirically
satisfies~\cref{eq:encoder_constraint}.
We discuss the detailed procedure in~\cref{sec:training}.

We compute a certificate for individual fairness in two steps.
First, to provide guarantees on the latent representation generated by encoder
$f_\theta$, we solve the optimization problem
\begin{equation*}
    \epsilon = \max_{x' \in S_\phi(x)} ||z - f_\theta(x')||_\infty.
\end{equation*}
Recall that the set $S_\phi(x)$ generally contains an infinite number of
individuals $x'$, and thus this optimization problem cannot be solved by simple
enumeration.
In~\cref{sec:certifying} we show how this optimization problem can be encoded
as a mixed-integer linear program (MILP) and solved using off-the-shelf MILP
solvers.
After obtaining $\epsilon$, we certify local robustness of the classifier
$h_\psi$ around $z = f_\theta(x)$ by proving (using MILP) that for each $z'$
where $||z' - z|| \leq \epsilon$, the classification results of $h_\psi(z')$
and $h_\psi(z)$ coincide.
Altogether, this implies the overall model $M = h_\psi \circ f_\theta$
satisfies individual fairness for $x$.
Finally, note that since the bounding box $\mathbb{B}\left(z, \epsilon\right)$
is a convex relaxation of the latent set of similar individuals
$f_\theta(S_\phi(x))$, the number of individuals for which we can obtain a
certificate is generally lower than the number of individuals
that actually satisfy~\cref{eq:individual-fairness}.

    \section{Related Work}
\label{sec:related-work}

We now discuss work most closely related to ours.

\paragraph{Learning fair representations}

There has been a long line of work on learning fair representations.
\citet{zemel2013learning} introduced a method to learn fair representations that
ensure group fairness and protect sensitive attributes.
Such representations, invariant to sensitive attributes, can also be learned
using variational autoencoders~\cite{louizos2016vae}, adversarial
learning~\cite{madras2018learning, edwards2016censoring}, or
disentanglement~\cite{creager2019flexibly}.
\citet{zemel2013learning} and \citet{madras2018learning} also consider the
problem of fair transfer learning, which we investigate in our work.
\citet{song2019controllable} used duality to unify some of the mentioned work
under the same framework.
\citet{mcnamara2019costs} derived theoretical guarantees for learning fair
representations.
Their guarantees require statistics of the data distribution and consist of
probabilistic bounds for individual and group fairness: for a new data point from
the same distribution, the constraint will hold with a certain probability.
In contrast, we obtain a certificate for a fixed data point, which ensures that
the fairness constraints holds (independent of the other data points).

Most work so far focuses on learning representations that satisfy statistical
notions of fairness, but there has also been some recent work on learning
individually fair representations.
These works learn fair representations with alternative definitions of
individual fairness based on Wasserstein distance~\cite{yurochkin2020training,
feng2019learning}, fairness graphs~\cite{lahoti2019operationalizing}, or
distance measures~\cite{lahoti2019ifair}.
A different line of work has investigated leaning the fairness metric from
data~\cite{yurochkin2020training, wang2019empirical, ilvento2020metric,
mukherjee2020two}.
In contrast, we define individual fairness via interpretable logical constraints.
Finally, recent works~\cite{garg2018a, pensia2020extracting, zhu2020learning}
studied the task of learning representations that are robust to (adversarial)
perturbations, \ie all similar individuals in our case, however not in the
context of fairness.
Many of the above methods for learning (individually) fair representations
employ nonlinear components~\cite{zemel2013learning, lahoti2019ifair},
graphs~\cite{lahoti2019operationalizing}, or sampling~\cite{louizos2016vae,
creager2019flexibly} and can thus not be efficiently certified, unlike the
neural networks that we consider in our work.

While we focus on learning fair representations, other lines of work have
investigated individual fairness in the context of clustering~\cite{jung2019a,
mahabadi2020individual}, causal inference~\cite{kusner2017counterfactual,
zhang2017a, madras2019fairness, chikahara2020learning}, composition of
individually fair classifiers~\cite{dwork2018fairness, dwork2020individual}, and
differential privacy (DP)~\cite{dwork2012fairness, jagielski2019differentially,
xu2019achieving}.
The close relationship between individual fairness and DP has been discussed in
previous work (see, \eg \cite{dwork2012fairness}).
However, DP crucially differs from our work in that it obtains a probabilistic
fairness guarantee, similar to \citet{mcnamara2019costs} mentioned above, whereas
we compute absolute fairness guarantees for every data point.
The most natural way to employ DP for a representation learning approach, like
LCIFR, would be to make the data producer model $f_{\theta}$ differentially
private for a neighborhood that encodes $S_\phi$, by adding noise inside the
computation of $f_{\theta}$.
If one can achieve DP for the neighborhood $S_\phi$ (a non-trivial challenge),
the data consumer model can then be seen as a post-processing step, which with
the right robustness certificate yields a probabilistic guarantee of
\cref{eq:individual-fairness}.

\paragraph{Certification of neural networks}

Certification of neural networks has become an effective way to prove that these
models are robust to adversarial perturbations.
Certification approaches are typically based on SMT
solving~\cite{katz2017reluplex}, abstract interpretation~\cite{gehr2018ai2},
mixed-integer linear programming~\cite{tjeng2017evaluating}, or linear
relaxations~\cite{zhang2018crown, singh2019deeppoly, singh2019krelu}.
Three concurrent works also investigate the certification of individual
fairness~\cite{urban2019perfectly, yeom2020individual, john2020verifying}.
However, these works try to certify a global individual fairness property, \ie
for a given classifier there exists an input which is not treated individually
fair, whereas we focus on local individual fairness, \ie for every concrete data
point we certify whether the model is individually fair or not.
Moreover, these works consider similarity notions that are less expressive than
and can be captured by our logical constraints.
Finally, they only consider certifying fairness of existing models, while we
also focus on learning fair representations.

In our work, we investigate modular certification.
For the data producer, we need to propagate the input shape through both logical
operators (\eg conjunctions and disjunctions) and the neural network.
While in our work, we use a MILP encoding, other approaches could also be
applied by crafting specialized convex relaxations.
For example, if our approach is applied to learn individually fair
representations of complex data such as images, where encoder networks are
usually larger than in tabular data that we consider here, one could leverage
the certification framework from \citet{singh2019krelu}.
On the data consumer side, any of the above approaches could be applied as they
are all designed to certify $\ell_\infty$-robustness which we consider in our
work.

    \section{Learning Individually Fair Representations}
\label{sec:training}

We now present our method for learning individually fair representations with
respect to the property $\phi$.
To illustrate our method, we consider the case where the regulator proposes the
similarity constraint:
\begin{equation*}
    \phi(x, x') := \bigwedge_{i \in \mathrm{Cat}\setminus\{\mathrm{race}\}} (x_{i} = x'_{i})
    \bigwedge_{j \in \mathrm{Num}} |x_{j} - x'_{j}| \leq \alpha.
\end{equation*}
According to $\phi$, individual $x'$ is considered similar to $x$ if: (i) all
categorical attributes except for race are equal to those of $x$, and (ii) all
numerical attributes (\eg income) of $x$ and $x'$ differ by at most $\alpha$.
Thus, under $\phi$, the similarity of individuals $x$ and $x'$ does not depend
on their respective races.
Note that since $\phi$ is binary $x$ and $x'$ are either considered similar or
not which is in line with the typical use-case in classification where two
individuals are either classified to the same label or not.
Moreover, such logical formulas (of reasonable size) are generally considered
humanly readable and are thus investigated in the interpretable machine
learning community (\eg for decision
trees~\cite{doshi-velez2018considerations}).


\paragraph{Enforcing individual fairness}
\label{subsec:training:enforcing}

To learn a representation that satisfies $\phi$, we build on the recent work
DL2~\cite{fischer2019dl2}.
Concretely, we aim to enforce the following constraint on the encoder
$f_\theta$ used by the data producer:
\begin{equation}
    \label{eq:dl2_encoder_constraint}
    \phi(x, x') \implies \| f_{\theta} \left( x \right) - f_{\theta} \left( x' \right)
    \|_{\infty} \leq \delta,
\end{equation}
where $\delta$ is a tunable constant, determined in agreement between the data
producer and the data consumer.
With DL2, this implication can be translated into a non-negative,
differentiable loss $\mathcal{L} \left( \phi \right)$ such that $\mathcal{L}
\left( \phi \right) \left( x, x' \right) = 0$ if and only if the implication is
satisfied.
Here, we denote $\omega \left( x, x' \right) := \| f_{\theta} \left( x \right)
- f_{\theta} \left( x' \right) \|_{\infty} \leq \delta$ and translate the
constraint in \cref{eq:dl2_encoder_constraint} as
\begin{equation*}
    \mathcal{L} \left( \phi \implies \omega \right)
    = \mathcal{L} \left( \neg \phi \lor \omega \right)
    = \mathcal{L} \left( \neg \phi \right) \cdot \mathcal{L} \left( \omega \right),
\end{equation*}
where negations are propagated through constraints via standard logic.
Moreover, we have
\begin{align*}
    \mathcal{L} \left( \omega \right) \left( x, x' \right)
    &= \mathcal{L} \left( \| f_{\theta} \left( x \right) - f_{\theta} \left( x' \right) \|_{\infty} \leq \delta \right) \\
    &= \max \left\{ \| f_{\theta} \left( x \right) - f_{\theta} \left( x' \right) \|_{\infty} - \delta, 0 \right\}.
\end{align*}
Similarly, conjunctions $\mathcal{L} \left( \phi' \land \phi'' \right)$ would
be translated as
$\mathcal{L} \left( \phi' \right) + \mathcal{L} \left( \phi'' \right)$, and we
refer interested readers to the original work~\cite{fischer2019dl2} for further
details on the translation.

Using this differentiable loss, the data producer can now approximate the
problem of finding an encoder $f_{\theta}$ that maximizes the probability that
the constraint $\phi \implies \omega$ is satisfied for all individuals via the
following min-max optimization problem (defined in two steps):
First, we find a counterexample
\begin{equation*}
    x^* = \argmin_{x' \in S_{\phi} \left( x \right)}
    \mathcal{L} \left( \neg \left( \phi \implies \omega \right) \right) \left( x, x' \right),
\end{equation*}
where $S_{\phi} \left( x \right) = \left\{ x' \in \mathbb{R}^n \mid \phi \left(
x, x' \right) \right\}$ denotes the set of all individuals similar to $x$ according to $\phi$.
Then, in the second step, we find the parameters $\theta$ that minimize the
constraint loss at $x^*$:
\begin{equation*}
    \argmin_{\theta} \mathbb{E}_{x \sim D} \left[ \mathcal{L} \left( \phi \implies \omega \right) \left( x, x^* \right) \right].
\end{equation*}
Note that in the outer loop, we are finding parameters $\theta$ that minimize
the loss of the original constraint from \cref{eq:dl2_encoder_constraint},
while in the inner loop, we are finding a counterexample $x^*$ by minimizing
the loss corresponding to the negation of this constraint.
We use Adam~\cite{kingma2015adam} for optimizing the outer problem.
For the inner minimization problem,~\citet{fischer2019dl2} further refine the
loss by excluding constraints that have closed-form analytical solutions, \eg
$\max \left\{ \left\| x - x' \right\|_{\infty} - \delta, 0 \right\}$ which can
be minimized by projecting $x'$ onto the $\ell_{\infty}$-ball of radius
$\delta$ around $x$.
The resulting objective is thus
\begin{equation*}
    x^* = \argmin_{x' \in \mathbb{C}} \mathcal{L} \left( \rho \right) \left( x, x' \right),
\end{equation*}
where $\mathbb{C}$ is the convex set and $\rho$ is
$\neg \left( \phi \implies \omega \right)$ without the respective constraints.
It has been shown~\cite{madry2018towards} that such an objective can be
efficiently solved with Projected Gradient Descent (PGD).

DL2 does not provide a meaningful translation for categorical constraints,
which are essential to fairness, and we derive a relaxation method for training
with categorical constraints in~\cref{sec:additional-training}.

\paragraph{Predictive utility of the representation}
\label{subsec:training:maintaining}

Recall that our method is modular in the sense that the data producer and the data
consumer models are learned separately. Thus, the data producer needs to ensure that the latent representation
remains informative for downstream applications (represented by the data
consumer model $h_\psi$).
To that end, the data producer additionally trains a classifier
$q \colon \mathbb{R}^k \rightarrow \mathbb{R}^o$ that tries to predict the
target label $y$ from the latent representation $z = f_\theta(x)$.
Thus, the data producer seeks to jointly train the encoder $f_{\theta}$ and
classifier $q$ to minimize the combined objective
\begin{equation}
    \label{eq:producer-objective}
    \argmin_{f_\theta, q} \mathbb{E}_{x, y} \left[
    \mathcal{L}_C \left( q \left( f_{\theta} \left( x \right) \right), y \right) + \gamma \mathcal{L}_F \left( x, f_\theta(x) \right)
    \right],
\end{equation}
where $\mathcal{L}_C$ is any suitable classification loss (\eg cross-entropy),
$\mathcal{L}_F$ is the fairness constraint loss obtained via DL2, and the
hyperparameter $\gamma$ balances the two objectives.
We empirically investigate impact of the loss balancing factor $\gamma$ on the
accuracy-fairness tradeoff in~\cref{sec:loss-balancing-factor}.

\paragraph{Training robust classifier $h_\psi$}
\label{subsec:training:robustconsumer}

We assume the encoder $f_\theta$ has been trained to maintain predictive
utility and satisfy~\cref{eq:dl2_encoder_constraint}.
Recall that, given this assumption, the data consumer who wants to ensure her
classifier $h_\psi$ is individually fair, only needs to ensure local robustness
of the classifier for perturbations up to $\delta$ in $l_\infty$-norm.
This is a standard problem in robust machine learning~\cite{ben2009robust} and
can be solved via min-max optimization, recently found to work well for neural
network models~\cite{madry2018towards}:
\begin{equation*}
    \min_\psi \mathbb{E}_{z \sim \mathcal{D}_z} \left[
    \max_{\pi \in \left[ \pm \delta\right]} \mathcal{L}_C \left( h_\psi(z + \pi), y \right)
    \right],
\end{equation*}
where $D_z$ is the latent distribution obtained by sampling from $\mathcal{D}$
and applying the encoder $f_\theta$, and $\mathcal{L}_C$ is a suitable
classification loss.
The optimization alternates between: (i) trying to find
$\pi \in \left[ \pm \delta \right]$ that maximizes
$\mathcal{L}_C \left( h_\psi(z + \pi), y \right)$, and (ii) updating $\psi$ to
minimize $\mathcal{L}_C \left( h_\psi(z + \pi), y \right)$ under such worst-case
perturbations $\pi$.
While the theoretical necessity of training for local robustness is clear, we
empirically investigate its effect on accuracy and certifiable fairness
in~\cref{sec:robust-training}.

    \section{Certifying Individual Fairness}
\label{sec:certifying}

In this section we discuss how the data consumer can compute a certificate of
individual fairness for its model $h_\psi$ trained on the latent representation
(as described in~\cref{subsec:training:robustconsumer} above).
We split this process into two steps: (i)~the data producer propagates a data
point $x$ through the encoder to obtain $z = f_\theta \left( x \right)$ and
computes the radius $\epsilon$ of the smallest $\ell_\infty$-ball around $z$
that contains the latent representations of all similar individuals
$f_\theta \left( S_\phi \left( x \right) \right)$, \ie
$f_\theta \left( S_\phi \left( x \right) \right) \subseteq \mathbb{B}_\infty \left( z, \epsilon \right)$,
and (ii)~the data consumer checks whether all points in the latent space that
differ by at most $\epsilon$ from $z$ are classified to the same label, \ie
$h_\psi \left( z \right) = h_\psi \left( z' \right)$ for all
$z' \in \mathbb{B}_\infty \left( z, \epsilon \right)$.
We now discuss both of these steps.

\subsection{Certifying Latent Similarity}
\label{subsec:certifying:latent}

To compute the minimum $\epsilon$ which ensures that $f_\theta \left( S_\phi
\left( x \right) \right) \subseteq \mathbb{B}_\infty \left( z, \epsilon
\right)$, the data producer models the set of similar individuals $S_\phi
\left( x \right)$ and the encoder $f_\theta$ as a mixed-integer linear program
(MILP).

\paragraph{Modeling $S_\phi$ as MILP}

We use an example to demonstrate the encoding of logical constraints with MILP.
Consider an individual $x$ that has two categorical features
$x_1 = \left[ 1, 0, \ldots, 0 \right]$ and
$x_2 = \left[ 0, \ldots, 0, 1 \right]$ and one
numerical feature $x_3$, with the following constraint for similarity:
\begin{equation*}
    \phi \left( x, x' \right) := \left( x_1 = x_1' \right) \land \left( | x_3 - x_3' | \leq \alpha \right).
\end{equation*}
Here $x$ is an individual from the test dataset and can be treated as constant,
while $x'$ is encoded using mixed-integer variables.
For every categorical feature $x_i'$ we introduce $k$ binary variables $v_i^l$
with $l = 1, \ldots, k$, where $k$ is the number of distinct values this
categorical feature can take.
For the fixed categorical feature $x_1'$, which is equal to $x_1$, we add the
constraints $v_1^1 = 1$ and $v_1^l = 0$ for $l = 2, \ldots, k$.
To model the free categorical feature $x_2'$ we add the constraint $\sum_{l}
v_2^l = 1$ thereby enforcing it to take on exactly one of $k$ potential values.
Finally, the numerical attribute $x_3'$ can be modeled by adding a
corresponding variable $v_3$ with the two constraints: $v_3 \geq x_3 - \alpha $
and $v_3 \leq x_3 + \alpha$.
It can be easily verified that our encoding of $S_\phi$ is exact.

Consider now a fairness constraint including disjunctions, \ie
$\phi := \phi_1 \lor \phi_2$.
To model such a disjunction we introduce two auxiliary binary variables $v_1$
and $v_2$ with the constraints $v_i = 1 \iff \phi_i \left( x, x' \right) = 1$
for $i = 1, 2$ and $v_1 + v_2 \geq 1$.

\paragraph{Handling general constraints}

The encodings demonstrated on these two examples can be applied for general
constraints $\phi$. A full formalization of our encoding is found in
Appendix~\ref{sec:full-encoding}.

\paragraph{Modeling $f_\theta$ as MILP}

To model the encoder we employ the method from~\citet{tjeng2017evaluating}
which is exact for neural networks with ReLU activations.
We recall that a ReLU performs $\max \left\{ x, 0 \right\}$ for some input $x$.
Given an upper and lower bound on $x$, \ie $x \in \left[ l, u \right]$ we can
encode the output of ReLU exactly via case distinction: (i) if $u \leq 0$ add a
variable with upper and lower bound $0$ to MILP, (ii) if $l \geq 0$ add a
variable with upper and lower bounds $u$ and $l$ respectively to MILP, and
(iii) if $l < 0 < u$, add a variable $v$ and a binary indicator $i$ to MILP in
addition to the following constraints:
\begin{gather*}
    0 \leq v \leq x \cdot i, \\
    x \leq v \leq x - l \cdot (1 - i), \\
    i = 1 \iff 0 \leq x.
\end{gather*}

Finally, given the MILP formulation of $S_\phi$ and $f_\theta$ we can compute
$\epsilon$ by solving the following $k$ MILP instances (where $k$ is the
dimension of the latent space):
\begin{equation*}
    \hat{\epsilon}_j = \max_{x' \in S_\phi(x)} |f_\theta^{\left( j \right)} \left( x \right) - f_\theta^{\left( j \right)} \left( x' \right)|.
\end{equation*}
We compute the final result as $\epsilon = \max\{\hat{\epsilon}_1,
\hat{\epsilon}_2, \ldots \hat{\epsilon}_k\}$.

\subsection{Certifying Local Robustness}
\label{subsec:certifying:local}

The data consumer obtains a point in latent space $z$ and a radius $\epsilon$.
To obtain a fairness certificate, the data consumer certifies that all points
in the latent space at $\ell_\infty$-distance at most $\epsilon$ from $z$ are
mapped to the same label as $z$.
This amounts to solving the following MILP optimization problem for each logit
$h^{( y' )}_\psi$ with label $y'$ different from the true label $y$:
\begin{equation*}
    \max_{z' \in \mathbb{B}_\infty(z, \epsilon)} h^{( y' )}_\psi(z') - h^{\left( y \right)}_\psi(z').
\end{equation*}
If the solution of the above optimization problem is less than zero for each
$y' \neq y$, then robustness of the classifier $h_\psi$ is provably
established.
Note that, the data consumer can employ same methods as the data producer to
encode the classifier as MILP~\cite{tjeng2017evaluating} and benefit from any
corresponding advancements in solving MILP instances in the context of neural
network certification, e.g.,~\cite{singh2019boosting}.

We now formalize our certificate, that allows the data consumer to prove
individual fairness of $M$, once given $z$ and $\epsilon$ by the data producer:

\begin{theorem}{(Individual fairness certificate)} \label{thm:certificate}
    Suppose $M = h_\psi \circ f_\theta$ with data point $x$ and similarity
    notion $\phi$. Furthermore, let $z = f_\theta(x)$,
    $S_\phi(x) = \{x' \in \mathbb{R}^n \mid \phi(x, x')\}$ and
    $\epsilon = \max_{x' \in S_\phi(x)} ||z - f_\theta(x')||_\infty$.
    If
    \begin{equation*}
        \max_{z' \in \mathbb{B}_\infty(z, \epsilon)}
        h^{( y' )}_\psi(z') - h^{\left( y \right)}_\psi(z') < 0
    \end{equation*}
    for all labels $y'$ different from the true label $y$, then for all
    $x' \in S_\phi(x)$ we have $M(x) = M(x')$.
\end{theorem}

\begin{proof}
    Provided in~\cref{sec:fairness-certificate}.
\end{proof}

    \section{Experimental Evaluation}
\label{sec:experiments}

We implement our method in a tool called LCIFR and present an extensive
experimental evaluation.
We consider a variety of different datasets --- Adult~\cite{dua2019uci},
Compas~\cite{angwin2016machine}, Crime~\cite{dua2019uci},
German~\cite{dua2019uci}, Health (\url{https://www.kaggle.com/c/hhp}), and Law
School~\cite{wightman1998lsac} --- which we describe in detail
in~\cref{sec:datasets}.
We perform the following preprocessing on all datasets: (i) normalize numerical
attributes to zero mean and unit variance, (ii) one-hot encode categorical
features, (iii) drop rows and columns with missing values, and (iv) split into
train, test and validation sets.
Although we only consider datasets with binary classification tasks, we note
that our method straightforwardly extends to the multiclass case.
We perform all experiments on a desktop PC using a single GeForce RTX 2080 Ti
GPU and 16-core Intel(R) Core(TM) i9-9900K CPU @ 3.60GHz.
We make all code, datasets and preprocessing pipelines publicly
available at \url{https://github.com/eth-sri/lcifr} to ensure reproducibility
of our results.

\paragraph{Experiment setup}

We model the encoder $f_\theta$ as a neural network, and we use logistic
regression as a classifier $h_\psi$.
We perform a grid search over model architectures and loss balancing factors
$\gamma$ which we evaluate on the validation set.
As a result, we consider $f_\theta$ with 1 hidden layer of 20 neurons (except
for Law School where we do not have a hidden layer) and a latent space of
dimension 20.
We fix $\gamma$ to 10 for Adult, Crime, and German, to 1 for Compas and Health,
and to 0.1 for Law School.
We provide a more detailed overview of the model architectures and hyperparameters
in~\cref{sec:experiment-setup}.

\paragraph{Fairness constraints}

We propose a range of different constraints for which we apply our method.
These constraints define the similarity between two individuals based on their
numerical attributes (\textsc{Noise}), categorical attributes (\textsc{Cat}),
or combinations thereof (\textsc{Cat + Noise}).
Furthermore, we consider more involved similarity notions based on disjunctions
(\textsc{Attribute}) and quantiles of certain attributes to counter
subordination between social
groups~\cite{lahoti2019operationalizing} (\textsc{Quantiles}).
A full formalization of our constraints is found in
Appendix~\ref{sec:constraints}.

\begin{table}
    \caption{
        Accuracy and certified individual fairness.
        We compare the accuracy and percentage of certified individuals with a
        baseline obtained from setting the loss balancing factor $\gamma = 0$.
        LCIFR produces a drastic increase in certified individuals while only
        incurring minor decrease in accuracy.
    }
    \label{tab:constraints}
    \begin{center}
        \begin{small}
            \begin{sc}
                \resizebox{0.7\columnwidth}{!}{ \begin{tabular}{cccccc}
    \toprule
    & & \multicolumn{2}{c}{Accuracy (\%)} & \multicolumn{2}{c}{Certified (\%)} \\
    Constraint & Dataset & Base & LCIFR & Base & LCIFR \\
    \midrule
    \multirow{6}{*}{Noise} & Adult & 83.0 & 81.4 & 59.0 & 97.8 \\
    & Compas & 65.8 & 63.4 & 32.1 & 79.0 \\
    & Crime & 84.4 & 83.1 & 7.4 & 66.9 \\
    & German & 76.5 & 74.0 & 71.0 & 97.5 \\
    & Health & 80.8 & 81.1 & 75.4 & 97.8 \\
    & Law School & 84.4 & 84.6 & 57.9 & 69.2 \\
    \midrule
    \multirow{6}{*}{Cat} & Adult & 83.3 & 83.1 & 79.9 & 100 \\
    & Compas & 65.6 & 66.3 & 90.9 & 100 \\
    & Crime & 84.4 & 83.9 & 78.3 & 100 \\
    & German & 76.0 & 75.5 & 88.5 & 100 \\
    & Health & 80.7 & 80.9 & 64.1 & 99.8 \\
    & Law School & 84.4 & 84.4 & 25.6 & 51.1 \\
    \midrule
    \multirow{6}{*}{Cat + Noise} & Adult & 83.3 & 81.3 & 47.5 & 97.6 \\
    & Compas & 65.6 & 63.7 & 30.9 & 75.6 \\
    & Crime & 84.4 & 81.5 & 6.2 & 63.3 \\
    & German & 76.0 & 70.0 & 68.0 & 95.5 \\
    & Health & 80.7 & 80.7 & 24.7 & 97.3 \\
    & Law School & 84.4 & 84.5 & 11.6 & 28.9 \\
    \midrule
    \multirow{3}{*}{Attribute} & Adult & 83.0 & 80.9 & 49.3 & 94.6 \\
    & German & 76.5 & 73.5 & 65.0 & 96.5 \\
    & Law School & 84.3 & 86.9 & 46.4 & 62.6 \\
    \midrule
    Quantiles & Law School & 84.2 & 84.2 & 56.5 & 76.9 \\
    \bottomrule
\end{tabular}
 }
            \end{sc}
        \end{small}
    \end{center}
\end{table}

\paragraph{Applying our method in practice}

We assume that the data regulator has defined the above constraints.
First, we act as the data producer and learn a representation that enforces
the individual fairness constraints using our method
from~\cref{subsec:training:enforcing}.
After training, we compute $\epsilon$ for every individual data point in the
test set and pass it to the data consumer along with the latent representation
of the entire dataset as described in~\cref{subsec:certifying:latent}.
Second, we act as data consumer and use our method
from~\cref{subsec:training:robustconsumer} to learn a locally-robust classifier
from the latent representation.
Finally, to obtain a certificate of individual fairness, we use $\epsilon$
to certify the classifier via our method from~\cref{subsec:certifying:local}.

In~\cref{tab:constraints} we compare the accuracy and percentage of certified
individuals (\ie the empirical approximation of a lower bound
on~\cref{eq:fair-risk}) with a baseline encoder and classifier obtained from
standard representation learning (\ie $\gamma = 0$).
We do not compare with other approaches for learning individually fair
representations since they either consider a different similarity metric or
employ nonlinear components that cannot be efficiently certified.
It can be observed that LCIFR drastically increases the percentage of certified
individuals across all constraints and datasets.
We would like to highlight the relatively low (albeit still significantly
higher than baseline) certification rate for the Law School dataset.
This is due to the relatively small loss balancing factor $\gamma = 0.1$ which
only weakly enforces the individual fairness constraint during training.
We report the following mean certification runtime per input, averaged
over all constraints: 0.29s on Adult, 0.35s on Compas, 1.23s on Crime, 0.28s on
German, 0.68s on Health, and 0.02s on Law School, showing that our method is
computationally efficient.
We show that our method scales to larger networks in~\cref{sec:scaling}.

\paragraph{Fair Transfer Learning}

We follow \citet{madras2018learning} to demonstrate that our method is
compatible with transferable representation learning.
We also consider the Health dataset, for which the original task is to predict
the Charlson Index.
To demonstrate transferability, we omit the primary condition group labels
from the set of features, and try to predict them from the latent
representation without explicitly optimizing for the task.
To that end, the data producer additionally learns a decoder
$g \left( z \right)$, which tries to predict the original attributes $x$ from
the latent representation, thereby not only retaining task-specific information
on the Charlson Index.
This amounts to adding a reconstruction loss $\mathcal{L}_R \left( x, g \left(
f_{\theta} \left( x \right) \right) \right)$ (\eg $\ell_2$) to the objective
in~\cref{eq:producer-objective}.
Assuming that our representations are in fact transferable, the data consumer
is now free to choose any classification objective.
We note that our certification method straightforwardly extends to all possible
prediction tasks allowing the data consumer to obtain fairness certificates
regardless of the objective.
Here, we let the data consumer train classifiers for both the original task and
to predict the 5 most common primary condition group labels.
We display the accuracy and percentage of certified data points on all tasks
in~\cref{tab:transfer}.
The table shows that our learned representation transfers well across tasks
while additionally providing provable individual fairness guarantees.

\begin{table}[t]
    \caption{
    Accuracy and percentage of certified individuals for transferable
    representation learning on Health dataset with \textsc{Cat + Noise}
    constraint.
    The transfer labels are omitted during training and the data producer
    objective is augmented with a reconstruction loss.
    This allows the data consumer to achieve high accuracies and
    certification rates across a variety of (potentially unknown) tasks.
    }
    \label{tab:transfer}
    \begin{center}
        \begin{small}
            \begin{sc}
                \resizebox{0.75\columnwidth}{!}{ \begin{tabular}{cccc}
    \toprule
    Task & Label & Accuracy (\%) & Certified (\%) \\
    \midrule
    Original & charlson index & 73.8 & 96.9 \\
    \midrule
    \multirow{5}{*}{Transfer} & msc2a3 & 73.7 & 86.1 \\
    & metab3 & 75.4 & 93.6 \\
    & arthspin & 75.4 & 93.7 \\
    & neument & 73.8 & 97.1 \\
    & respr4 & 72.4 & 98.4 \\
    \bottomrule
\end{tabular}
 }
            \end{sc}
        \end{small}
    \end{center}
\end{table}

    \section{Conclusion}
\label{sec:conclusion}

We introduced a novel end-to-end framework for learning representations with
provable certificates of individual fairness.
We demonstrated that our method is compatible with existing notions of fairness,
such as transfer learning.
Our evaluation across different datasets and fairness constraints demonstrates
the practical effectiveness of our method.

    \section*{Broader Impact}

Methods that learn from data can potentially produce unfair outcomes by
reinforcing human biases or discriminating amongst specific groups.
We illustrate how our method can be employed to address these issues with an
example due to~\citet{cisse2019fairness}.
Consider a company with several teams working with the same data to build
models.
Although the individual teams may not care about fairness of their models, the
company needs to comply with ethical or legal requirements.
In this setting, our framework enables the company to obtain such certificates
from every team in a minimally invasive and modular fashion without compromising
downstream utility.

Although individual fairness is a desirable property, it is far from sufficient
to provide any ethical guarantees.
For example, treating all individuals similarly badly does not conflict with
individual fairness.
Our method thus depends on the assumption that all involved parties act
reasonably.
That is, the data regulator needs to take all ethical aspects and future
societal consequences into consideration when designing the similarity property.
However, even a diligent data regulator may unconsciously encode biases in the
similarity measure.
Moreover, our approach breaks down with an adversarial data producer that either
explicitly learns a discriminatory representation or simply fails to respect the
defined similarity notion.
Finally, the case where the data consumer acts adversarially has been
investigated in previous work~\cite{madras2018learning} and can be mitigated to
some extent.

    \begin{ack}
    We would like to thank Anna Chmurovič for her help with initial
    investigations on combining fairness and differentiable logic during her ETH
    Student Summer Research Fellowship.
    We also thank the anonymous reviewers for their insightful comments and
    suggestions.
\end{ack}

    \message{^^JLASTBODYPAGE \thepage^^J}

    \bibliography{references}

\begin{thebibliography}{59}
\providecommand{\natexlab}[1]{#1}
\providecommand{\url}[1]{\texttt{#1}}
\expandafter\ifx\csname urlstyle\endcsname\relax
  \providecommand{\doi}[1]{doi: #1}\else
  \providecommand{\doi}{doi: \begingroup \urlstyle{rm}\Url}\fi

\bibitem[Brennan et~al.(2009)Brennan, Dieterich, and Ehret]{brennan2009compas}
Tim Brennan, William Dieterich, and Beate Ehret.
\newblock Evaluating the predictive validity of the compas risk and needs
  assessment system.
\newblock \emph{Criminal Justice and Behavior}, 2009.

\bibitem[Datta et~al.(2015)Datta, Tschantz, and Datta]{datta2015automated}
Amit Datta, Michael~Carl Tschantz, and Anupam Datta.
\newblock Automated experiments on ad privacy settings.
\newblock \emph{Proceedings on Privacy Enhancing Technologies}, 2015.

\bibitem[Khandani et~al.(2010)Khandani, Kim, and Lo]{khandani2010consumer}
Amir~E Khandani, Adlar~J Kim, and Andrew~W Lo.
\newblock Consumer credit-risk models via machine-learning algorithms.
\newblock \emph{Journal of Banking \& Finance}, 2010.

\bibitem[Sweeney(2013)]{sweeney2013discrimination}
Latanya Sweeney.
\newblock Discrimination in online ad delivery.
\newblock \emph{Queue}, 2013.

\bibitem[Bolukbasi et~al.(2016)Bolukbasi, Chang, Zou, Saligrama, and
  Kalai]{bolukbasi2016man}
Tolga Bolukbasi, Kai{-}Wei Chang, James~Y. Zou, Venkatesh Saligrama, and
  Adam~Tauman Kalai.
\newblock Man is to computer programmer as woman is to homemaker? debiasing
  word embeddings.
\newblock In \emph{Advances in Neural Information Processing Systems 29}, 2016.

\bibitem[Barocas and Selbst(2016)]{barocas2016big}
Solon Barocas and Andrew~D. Selbst.
\newblock Big data's disparate impact.
\newblock \emph{California Law Review}, 2016.

\bibitem[Cisse and Koyejo(2019)]{cisse2019fairness}
Moustapha Cisse and Sanmi Koyejo.
\newblock Fairness and representation learning.
\newblock NeurIPS Invited Talk 2019, 2019.

\bibitem[Zemel et~al.(2013)Zemel, Wu, Swersky, Pitassi, and
  Dwork]{zemel2013learning}
Richard~S. Zemel, Yu~Wu, Kevin Swersky, Toniann Pitassi, and Cynthia Dwork.
\newblock Learning fair representations.
\newblock In \emph{Proceedings of the 30th International Conference on Machine
  Learning}, 2013.

\bibitem[Madras et~al.(2018)Madras, Creager, Pitassi, and
  Zemel]{madras2018learning}
David Madras, Elliot Creager, Toniann Pitassi, and Richard~S. Zemel.
\newblock Learning adversarially fair and transferable representations.
\newblock In \emph{Proceedings of the 35th International Conference on Machine
  Learning}, 2018.

\bibitem[McNamara et~al.(2017)McNamara, Ong, and
  Williamson]{mcnamara2017provably}
Daniel McNamara, Cheng~Soon Ong, and Robert~C Williamson.
\newblock Provably fair representations.
\newblock \emph{arXiv preprint arXiv:1710.04394}, 2017.

\bibitem[Chouldechova and Roth(2018)]{chouldechova2018frontiers}
Alexandra Chouldechova and Aaron Roth.
\newblock The frontiers of fairness in machine learning.
\newblock \emph{arXiv preprint arXiv:1810.08810}, 2018.

\bibitem[Dwork et~al.(2012)Dwork, Hardt, Pitassi, Reingold, and
  Zemel]{dwork2012fairness}
Cynthia Dwork, Moritz Hardt, Toniann Pitassi, Omer Reingold, and Richard Zemel.
\newblock Fairness through awareness.
\newblock In \emph{Proceedings of the 3rd Innovations in Theoretical Computer
  Science Conference}, 2012.

\bibitem[Hardt et~al.(2016)Hardt, Price, and Srebro]{hardt2016equality}
Moritz Hardt, Eric Price, and Nati Srebro.
\newblock Equality of opportunity in supervised learning.
\newblock In \emph{Advances in Neural Information Processing Systems 29}, 2016.

\bibitem[Kearns et~al.(2018)Kearns, Neel, Roth, and Wu]{kearns2018preventing}
Michael~J. Kearns, Seth Neel, Aaron Roth, and Zhiwei~Steven Wu.
\newblock Preventing fairness gerrymandering: Auditing and learning for
  subgroup fairness.
\newblock In \emph{Proceedings of the 35th International Conference on Machine
  Learning}, 2018.

\bibitem[Fischer et~al.(2019)Fischer, Balunovic, Drachsler{-}Cohen, Gehr,
  Zhang, and Vechev]{fischer2019dl2}
Marc Fischer, Mislav Balunovic, Dana Drachsler{-}Cohen, Timon Gehr, Ce~Zhang,
  and Martin~T. Vechev.
\newblock {DL2:} training and querying neural networks with logic.
\newblock In \emph{Proceedings of the 36th International Conference on Machine
  Learning}, 2019.

\bibitem[Tjeng et~al.(2019)Tjeng, Xiao, and Tedrake]{tjeng2017evaluating}
Vincent Tjeng, Kai~Y. Xiao, and Russ Tedrake.
\newblock Evaluating robustness of neural networks with mixed integer
  programming.
\newblock In \emph{7th International Conference on Learning Representations},
  2019.

\bibitem[Lahoti et~al.(2019{\natexlab{a}})Lahoti, Gummadi, and
  Weikum]{lahoti2019ifair}
Preethi Lahoti, Krishna~P. Gummadi, and Gerhard Weikum.
\newblock ifair: Learning individually fair data representations for
  algorithmic decision making.
\newblock In \emph{35th {IEEE} International Conference on Data Engineering},
  2019{\natexlab{a}}.

\bibitem[Lahoti et~al.(2019{\natexlab{b}})Lahoti, Gummadi, and
  Weikum]{lahoti2019operationalizing}
Preethi Lahoti, Krishna~P. Gummadi, and Gerhard Weikum.
\newblock Operationalizing individual fairness with pairwise fair
  representations.
\newblock \emph{Proc. {VLDB} Endow.}, 2019{\natexlab{b}}.

\bibitem[Yurochkin et~al.(2020)Yurochkin, Bower, and
  Sun]{yurochkin2020training}
Mikhail Yurochkin, Amanda Bower, and Yuekai Sun.
\newblock Training individually fair {ML} models with sensitive subspace
  robustness.
\newblock In \emph{8th International Conference on Learning Representations},
  2020.

\bibitem[Wang et~al.(2019)Wang, Grgic-Hlaca, Lahoti, Gummadi, and
  Weller]{wang2019empirical}
Hanchen Wang, Nina Grgic-Hlaca, Preethi Lahoti, Krishna~P. Gummadi, and Adrian
  Weller.
\newblock An empirical study on learning fairness metrics for compas data with
  human supervision.
\newblock \emph{arXiv preprint arXiv:1910.10255}, 2019.

\bibitem[Ilvento(2020)]{ilvento2020metric}
Christina Ilvento.
\newblock {Metric Learning for Individual Fairness}.
\newblock In \emph{1st Symposium on Foundations of Responsible Computing},
  2020.

\bibitem[Louizos et~al.(2016)Louizos, Swersky, Li, Welling, and
  Zemel]{louizos2016vae}
Christos Louizos, Kevin Swersky, Yujia Li, Max Welling, and Richard~S. Zemel.
\newblock The variational fair autoencoder.
\newblock In \emph{4th International Conference on Learning Representations},
  2016.

\bibitem[Edwards and Storkey(2016)]{edwards2016censoring}
Harrison Edwards and Amos~J. Storkey.
\newblock Censoring representations with an adversary.
\newblock In \emph{4th International Conference on Learning Representations},
  2016.

\bibitem[Creager et~al.(2019)Creager, Madras, Jacobsen, Weis, Swersky, Pitassi,
  and Zemel]{creager2019flexibly}
Elliot Creager, David Madras, J{\"{o}}rn{-}Henrik Jacobsen, Marissa~A. Weis,
  Kevin Swersky, Toniann Pitassi, and Richard~S. Zemel.
\newblock Flexibly fair representation learning by disentanglement.
\newblock In \emph{Proceedings of the 36th International Conference on Machine
  Learning}, 2019.

\bibitem[Song et~al.(2019)Song, Kalluri, Grover, Zhao, and
  Ermon]{song2019controllable}
Jiaming Song, Pratyusha Kalluri, Aditya Grover, Shengjia Zhao, and Stefano
  Ermon.
\newblock Learning controllable fair representations.
\newblock In \emph{The 22nd International Conference on Artificial Intelligence
  and Statistics}, 2019.

\bibitem[McNamara et~al.(2019)McNamara, Ong, and Williamson]{mcnamara2019costs}
Daniel McNamara, Cheng~Soon Ong, and Robert~C. Williamson.
\newblock Costs and benefits of fair representation learning.
\newblock In \emph{Proceedings of the 2019 AAAI/ACM Conference on AI, Ethics,
  and Society}, 2019.

\bibitem[Feng et~al.(2019)Feng, Yang, Lyu, Tan, Sun, and
  Wang]{feng2019learning}
Rui Feng, Yang Yang, Yuehan Lyu, Chenhao Tan, Yizhou Sun, and Chunping Wang.
\newblock Learning fair representations via an adversarial framework.
\newblock \emph{arXiv preprint arXiv:1904.13341}, 2019.

\bibitem[Mukherjee et~al.(2020)Mukherjee, Yurochkin, Banerjee, and
  Sun]{mukherjee2020two}
Debarghya Mukherjee, Mikhail Yurochkin, Moulinath Banerjee, and Yuekai Sun.
\newblock Two simple ways to learn individual fairness metrics from data.
\newblock \emph{arXiv preprint arXiv:2006.11439}, 2020.

\bibitem[Garg et~al.(2018)Garg, Sharan, Zhang, and Valiant]{garg2018a}
Shivam Garg, Vatsal Sharan, Brian~Hu Zhang, and Gregory Valiant.
\newblock A spectral view of adversarially robust features.
\newblock In \emph{Advances in Neural Information Processing Systems 31}, 2018.

\bibitem[{Pensia} et~al.(2020){Pensia}, {Jog}, and {Loh}]{pensia2020extracting}
A.~{Pensia}, V.~{Jog}, and P.~{Loh}.
\newblock Extracting robust and accurate features via a robust information
  bottleneck.
\newblock \emph{IEEE Journal on Selected Areas in Information Theory}, 2020.

\bibitem[Zhu et~al.(2020)Zhu, Zhang, and Evans]{zhu2020learning}
Sicheng Zhu, Xiao Zhang, and David Evans.
\newblock Learning adversarially robust representations via worst-case mutual
  information maximization.
\newblock \emph{arXiv preprint arXiv:2002.11798}, 2020.

\bibitem[Jung et~al.(2019)Jung, Kannan, and Lutz]{jung2019a}
Christopher Jung, Sampath Kannan, and Neil Lutz.
\newblock A center in your neighborhood: Fairness in facility location.
\newblock \emph{arXiv preprint arXiv:1908.09041}, 2019.

\bibitem[Mahabadi and Vakilian(2020)]{mahabadi2020individual}
Sepideh Mahabadi and Ali Vakilian.
\newblock Individual fairness for $k$-clustering.
\newblock \emph{arXiv preprint arXiv:2002.06742}, 2020.

\bibitem[Kusner et~al.(2017)Kusner, Loftus, Russell, and
  Silva]{kusner2017counterfactual}
Matt~J. Kusner, Joshua~R. Loftus, Chris Russell, and Ricardo Silva.
\newblock Counterfactual fairness.
\newblock In \emph{Advances in Neural Information Processing Systems 30}, 2017.

\bibitem[Zhang et~al.(2017)Zhang, Wu, and Wu]{zhang2017a}
Lu~Zhang, Yongkai Wu, and Xintao Wu.
\newblock A causal framework for discovering and removing direct and indirect
  discrimination.
\newblock In \emph{Proceedings of the Twenty-Sixth International Joint
  Conference on Artificial Intelligence}, 2017.

\bibitem[Madras et~al.(2019)Madras, Creager, Pitassi, and
  Zemel]{madras2019fairness}
David Madras, Elliot Creager, Toniann Pitassi, and Richard~S. Zemel.
\newblock Fairness through causal awareness: Learning causal latent-variable
  models for biased data.
\newblock In \emph{Proceedings of the Conference on Fairness, Accountability,
  and Transparency}, 2019.

\bibitem[Chikahara et~al.(2020)Chikahara, Sakaue, Fujino, and
  Kashima]{chikahara2020learning}
Yoichi Chikahara, Shinsaku Sakaue, Akinori Fujino, and Hisashi Kashima.
\newblock Learning individually fair classifier with path-specific
  causal-effect constraint.
\newblock \emph{arXiv preprint arXiv:2002.06746}, 2020.

\bibitem[Dwork and Ilvento(2018)]{dwork2018fairness}
Cynthia Dwork and Christina Ilvento.
\newblock {Fairness Under Composition}.
\newblock In \emph{10th Innovations in Theoretical Computer Science
  Conference}, 2018.

\bibitem[Dwork et~al.(2020)Dwork, Ilvento, and Jagadeesan]{dwork2020individual}
Cynthia Dwork, Christina Ilvento, and Meena Jagadeesan.
\newblock {Individual Fairness in Pipelines}.
\newblock In \emph{1st Symposium on Foundations of Responsible Computing},
  2020.

\bibitem[Jagielski et~al.(2019)Jagielski, Kearns, Mao, Oprea, Roth,
  Sharifi{-}Malvajerdi, and Ullman]{jagielski2019differentially}
Matthew Jagielski, Michael~J. Kearns, Jieming Mao, Alina Oprea, Aaron Roth,
  Saeed Sharifi{-}Malvajerdi, and Jonathan Ullman.
\newblock Differentially private fair learning.
\newblock In \emph{Proceedings of the 36th International Conference on Machine
  Learning}, 2019.

\bibitem[Xu et~al.(2019)Xu, Yuan, and Wu]{xu2019achieving}
Depeng Xu, Shuhan Yuan, and Xintao Wu.
\newblock Achieving differential privacy and fairness in logistic regression.
\newblock In \emph{Companion of The 2019 World Wide Web Conference}, 2019.

\bibitem[Katz et~al.(2017)Katz, Barrett, Dill, Julian, and
  Kochenderfer]{katz2017reluplex}
Guy Katz, Clark~W. Barrett, David~L. Dill, Kyle Julian, and Mykel~J.
  Kochenderfer.
\newblock Reluplex: An efficient {SMT} solver for verifying deep neural
  networks.
\newblock In \emph{Computer Aided Verification - 29th International
  Conference}, 2017.

\bibitem[Gehr et~al.(2018)Gehr, Mirman, Drachsler{-}Cohen, Tsankov, Chaudhuri,
  and Vechev]{gehr2018ai2}
Timon Gehr, Matthew Mirman, Dana Drachsler{-}Cohen, Petar Tsankov, Swarat
  Chaudhuri, and Martin~T. Vechev.
\newblock {AI2:} safety and robustness certification of neural networks with
  abstract interpretation.
\newblock In \emph{2018 {IEEE} Symposium on Security and Privacy}, 2018.

\bibitem[Zhang et~al.(2018)Zhang, Weng, Chen, Hsieh, and
  Daniel]{zhang2018crown}
Huan Zhang, Tsui{-}Wei Weng, Pin{-}Yu Chen, Cho{-}Jui Hsieh, and Luca Daniel.
\newblock Efficient neural network robustness certification with general
  activation functions.
\newblock In \emph{Advances in Neural Information Processing Systems 31}, 2018.

\bibitem[Singh et~al.(2019{\natexlab{a}})Singh, Gehr, P{\"{u}}schel, and
  Vechev]{singh2019deeppoly}
Gagandeep Singh, Timon Gehr, Markus P{\"{u}}schel, and Martin~T. Vechev.
\newblock An abstract domain for certifying neural networks.
\newblock \emph{Proc. {ACM} Program. Lang.}, 2019{\natexlab{a}}.

\bibitem[Singh et~al.(2019{\natexlab{b}})Singh, Ganvir, P{\"{u}}schel, and
  Vechev]{singh2019krelu}
Gagandeep Singh, Rupanshu Ganvir, Markus P{\"{u}}schel, and Martin~T. Vechev.
\newblock Beyond the single neuron convex barrier for neural network
  certification.
\newblock In \emph{Advances in Neural Information Processing Systems 32},
  2019{\natexlab{b}}.

\bibitem[Urban et~al.(2019)Urban, Christakis, W{\"u}stholz, and
  Zhang]{urban2019perfectly}
Caterina Urban, Maria Christakis, Valentin W{\"u}stholz, and Fuyuan Zhang.
\newblock Perfectly parallel fairness certification of neural networks.
\newblock \emph{arXiv preprint arXiv:1912.02499}, 2019.

\bibitem[Yeom and Fredrikson(2020)]{yeom2020individual}
Samuel Yeom and Matt Fredrikson.
\newblock Individual fairness revisited: Transferring techniques from
  adversarial robustness.
\newblock In \emph{Proceedings of the Twenty-Ninth International Joint
  Conference on Artificial Intelligence}, 2020.

\bibitem[John et~al.(2020)John, Vijaykeerthy, and Saha]{john2020verifying}
Philips~George John, Deepak Vijaykeerthy, and Diptikalyan Saha.
\newblock Verifying individual fairness in machine learning models.
\newblock In \emph{Proceedings of the Thirty-Sixth Conference on Uncertainty in
  Artificial Intelligence}, 2020.

\bibitem[Doshi-Velez and Kim(2018)]{doshi-velez2018considerations}
Finale Doshi-Velez and Been Kim.
\newblock \emph{Considerations for Evaluation and Generalization in
  Interpretable Machine Learning}.
\newblock Springer International Publishing, 2018.

\bibitem[Kingma and Ba(2015)]{kingma2015adam}
Diederik~P. Kingma and Jimmy Ba.
\newblock Adam: {A} method for stochastic optimization.
\newblock In \emph{3rd International Conference on Learning Representations},
  2015.

\bibitem[Madry et~al.(2018)Madry, Makelov, Schmidt, Tsipras, and
  Vladu]{madry2018towards}
Aleksander Madry, Aleksandar Makelov, Ludwig Schmidt, Dimitris Tsipras, and
  Adrian Vladu.
\newblock Towards deep learning models resistant to adversarial attacks.
\newblock In \emph{6th International Conference on Learning Representations},
  2018.

\bibitem[Ben-Tal et~al.(2009)Ben-Tal, El~Ghaoui, and Nemirovski]{ben2009robust}
Aharon Ben-Tal, Laurent El~Ghaoui, and Arkadi Nemirovski.
\newblock \emph{Robust optimization}.
\newblock Princeton University Press, 2009.

\bibitem[Singh et~al.(2019{\natexlab{c}})Singh, Gehr, P{\"{u}}schel, and
  Vechev]{singh2019boosting}
Gagandeep Singh, Timon Gehr, Markus P{\"{u}}schel, and Martin~T. Vechev.
\newblock Boosting robustness certification of neural networks.
\newblock In \emph{7th International Conference on Learning Representations},
  2019{\natexlab{c}}.

\bibitem[Dua and Graff(2017)]{dua2019uci}
Dheeru Dua and Casey Graff.
\newblock {UCI} machine learning repository, 2017.

\bibitem[Angwin et~al.(2016)Angwin, Larson, Mattu, and
  Kirchner]{angwin2016machine}
Julia Angwin, Jeff Larson, Surya Mattu, and Lauren Kirchner.
\newblock Machine bias, 2016.

\bibitem[Wightman(2017)]{wightman1998lsac}
F.~Linda Wightman.
\newblock {LSAC} national longitudinal bar passage study, 2017.

\bibitem[Paszke et~al.(2019)Paszke, Gross, Massa, Lerer, Bradbury, Chanan,
  Killeen, Lin, Gimelshein, Antiga, Desmaison, Kopf, Yang, DeVito, Raison,
  Tejani, Chilamkurthy, Steiner, Fang, Bai, and Chintala]{paszke2019pytorch}
Adam Paszke, Sam Gross, Francisco Massa, Adam Lerer, James Bradbury, Gregory
  Chanan, Trevor Killeen, Zeming Lin, Natalia Gimelshein, Luca Antiga, Alban
  Desmaison, Andreas Kopf, Edward Yang, Zachary DeVito, Martin Raison, Alykhan
  Tejani, Sasank Chilamkurthy, Benoit Steiner, Lu~Fang, Junjie Bai, and Soumith
  Chintala.
\newblock Pytorch: An imperative style, high-performance deep learning library.
\newblock In \emph{Advances in Neural Information Processing Systems 32}, 2019.

\bibitem[Ehlers(2017)]{ehlers2017formal}
R{\"{u}}diger Ehlers.
\newblock Formal verification of piece-wise linear feed-forward neural
  networks.
\newblock In \emph{Automated Technology for Verification and Analysis - 15th
  International Symposium}, 2017.

\end{thebibliography}
    \bibliographystyle{unsrtnat}

    \message{^^JLASTREFERENCESPAGE \thepage^^J}


    \ifbool{includeappendix}{%
    \clearpage
    \appendix
    \section{Training with Categorical Constraints}
\label{sec:additional-training}

A key challenge arising in~\cref{sec:training} is that DL2 does not support
logical formulas $\phi$ involving categorical constraints, which are critical to
the fairness context.
To illustrate this problem, we recall the example similarity constraint
\begin{equation*}
    \phi(x, x') := \bigwedge_{i \in \mathrm{Cat}\setminus\{\mathrm{race}\}} (x_{i} = x'_{i})
    \bigwedge_{j \in \mathrm{Num}} |x_{j} - x'_{j}| \leq \alpha.
\end{equation*}
As mentioned in~\cref{sec:training}, numerical attribute constraints of the form
$| x_j - x_j' | \leq \alpha$ can be solved efficiently by projecting $x_j'$ onto
$\left[ x_j - \alpha, x_j + \alpha \right]$.
Unfortunately, this does not directly extend to categorical constraints.
To see this, consider another constraint that considers two individuals $x$ and
$x'$ similar irrespective of their race.
Further, consider an individual $x$ with only one (categorical) attribute,
namely $x = \left[ \mathrm{race}_1 \right]$, and $r$ distinct races.
After a one-hot encoding, the features of $x$ are
$\left[ 1, 0, \ldots, 0 \right]$.
Now, one could try to translate the constraint as
$| x_{k} - x_{k}' | \leq \alpha$ for all $k = 1, \ldots, r$.
However, choosing \eg $\alpha = 0.3$ would only allow for $x'$ with features of
the form $\left[ 0.7, 0.3, 0, \ldots, 0 \right]$ which still represent the same
race when considering the maximum element.
Thus, this translation would not consider individuals with different races as
similar.
At the same time, choosing a larger $\alpha$, \eg $\alpha= 0.9$, would yield a
translation which considers an individual $x'$ with features
$\left[ 0.9, 0.9, \ldots, 0.9 \right]$ similar to $x$.
Clearly, this does not provide a meaningful relaxation of the categorical
constraint.

To overcome this problem, we relax the categorical constraint to
$x_{k}' \in \left[ 0, 1 \right]$ and normalize the sum over all possible races
as $\sum_k x_{k}' = 1$ with every projection step, thus ensuring a meaningful
feature vector.
Moreover, it can be easily seen that our translation allows $x'$ to take on any
race value irrespective of the race of $x$.
We note that although our relaxation can produce features with fractional
values, \eg $\left[ 0, 0.2, 0.3, 0, \ldots, 0.5 \right]$, we found it works well
in practice.

\section{Loss Balancing Factor $\gamma$}
\label{sec:loss-balancing-factor}

Here, we investigate the impact of the balance parameter $\gamma$ from
\cref{eq:producer-objective} on the accuracy-fairness tradeoff.
To that end, we compare accuracy and certified individual fairness for different
loss balancing factors for the \textsc{Cat + Noise} constraint on the
\textsc{Crime} dataset in \cref{tab:fairness-accuracy-tradeoff}.
We observe that increasing $\gamma$ up to 10 yields significant fairness gains
while keeping the accuracy roughly constant.
For larger values of $\gamma$, the fairness constraint dominates the loss and
causes the classifier to resort to majority class prediction, which is perfectly
fair.
Note that our method can increase both accuracy, albeit only by a small amount,
and fairness for certain values of $\gamma$ (\eg $\gamma = 2$).
We conjecture that this effect is due to randomness in the training procedure
and sufficient model capacity for simultaneous accuracy and fairness for $\gamma
\leq 5$.
As we observed the same trend on all datasets, we recommend data producers who
want to apply LCIFR in practice to increase $\gamma$ up to the point where the
downstream validation accuracy drops below their requirements.

\begin{table}
    \caption{
        Accuracy and certified individual fairness for the \textsc{Cat + Noise}
        constraint on the \textsc{Crime} dataset for different loss balancing
        factors $\gamma$.
        Compared to the baseline $\gamma = 0$, our method ($\gamma \neq 0$)
        incurs minimal changes in accuracy while significantly increasing the
        percentage of certified individual fairness for a wide range of
        $\gamma$.
    }
    \begin{center}
        \begin{small}
            \begin{sc}
                \resizebox{\columnwidth}{!}{ \begin{tabular}{ccccccccccc}
    \toprule
    $\gamma$ & 0 & 0.1 & 0.2 & 0.5 & 1 & 2 & 5 & 10 & 20 & 50 \\
    \midrule
    accuracy (\%) & 84.36 & 84.62 & 84.87 & 84.36 & 84.10 & 84.62 & 84.36 & 81.79 & 50.77 & 50.77 \\
    certified (\%) & 6.15 & 9.23 & 12.05 & 18.46 & 33.08 & 52.31 & 61.28 & 62.82 & 100 & 100 \\
    \bottomrule
\end{tabular}

 }
            \end{sc}
        \end{small}
    \end{center}
    \label{tab:fairness-accuracy-tradeoff}
\end{table}

\section{Robust Training}
\label{sec:robust-training}

Here, we investigate the necessity and impact of the robust training employed by the
data consumer as outlined in~\cref{sec:training}.
We recall that the data consumer obtains the latent representation
$z = f_\theta(x)$ for every data point $x$ from the data producer.
Assuming that the latent representation was generated by an encoder $f_\theta$
trained to maintain predictive utility and
satisfy~\cref{eq:dl2_encoder_constraint}, the data consumer only needs to ensure
local robustness of her classifier $h_\psi$ for perturbations up to $\delta$
in $\ell_\infty$-norm to obtain an individually fair classifier $h_\psi$.
However, the data consumer may be hestitant to apply robust training methods due
to potentially negative impacts on accuracy or may not care about fairness at all.

We first study the case where the data consumer employs logistic regression
for $h_\psi$ as in~\cref{sec:experiments}.
We consider the \textsc{Cat + Noise} constraint and run LCIFR on all five
datasets for different loss balancing factors
$\gamma \in \{0.001, 0.01, 0.1, 1, 10\}$ both with and without adversarial
training for $h_\psi$.
Across all datasets and values of $\gamma$ the largest increase in certification
for adversarial training is roughly 7\%, with a simultaneous accuracy drop of
0.5\%, and the largest accuracy drop is roughly 1\%, with a simultaneous
increase in certification of 2.9\%.
This rather limited impact of adversarial training on both accuracy and
certifiable individual fairness for logistic regression is to be expected due to
the smoothness of the decision boundary.
However, for a more complex classifier, such as a feedforward neural network
with 2 hidden layers of 20 nodes each, adversarial training doubles the
certification rate from 34\% to 70.8\%, while decreasing accuracy only by 1.6\%
for the \textsc{Cat + Noise} constraint on the \textsc{Health} datasets with
$\gamma = 1$.

\section{Full Encoding}
\label{sec:full-encoding}

Here, we present our fairness constraint language and show how to encode
constraints as a mixed-inter linear program (MILP).
We closely follow \citet{fischer2019dl2}.

\paragraph{Logical language}
We recall that our framework allows the data regulator to define notions of
similarity via a logical constraint $\phi$.
Our language of logical constraints consists of boolean combinations of
comparisons between terms where each term $t$ is a linear function over a
data point $x$.
We note that although \citet{fischer2019dl2} support terms with real-valued
functions, we only consider linear functions since nonlinear constraints,
\eg $x^2 < 3$, cannot be encoded exactly as MILP.
Unlike \citet{fischer2019dl2}, our constraint language also supports
constraints on categorical features.
To form comparison constraints, two terms $t$ and $t'$ can be combined as
$t = t'$, $t \leq t'$, $t \neq t'$, and $t < t'$.
Finally, a logical constraint $\phi$ is either a comparison constraint, a
negation $\lnot \phi'$ of a constraint $\phi'$, or a conjunction
$\phi' \land \phi''$ or disjunction $\phi' \lor \phi''$ of two constraints
$\phi'$ and $\phi''$.

\paragraph{Encoding as MILP}
Given an individual $x$ and a logical constraint $\phi$ capturing some
notion of similarity, the data producer needs to compute the radius $\epsilon$
of the smallest $\ell_\infty$-ball around the latent representation
$z = f_\theta\left(x\right)$ that contains the latent representations of all
similar individuals $f_\theta \left( S_\phi \left( x \right) \right)$,
\ie
$\argmin_\epsilon f_\theta \left( S_\phi \left( x \right) \right) \subseteq
\mathcal{B}_\infty \left( z, \epsilon \right)$.
To that end, the data producer is required to encode $S_\phi \left( x \right)$
as a MILP which can be performed in a recursive manner.

The individual $x$ belongs to the test dataset and can thus be treated as a
constant.
To model $S_\phi \left( x \right)$, we encode a similar individual $x'$ by
considering numerical and categorical features separately.
For all numerical features we add a real-valued variable $v_i$ to the MILP.
For all categorical features we add $k_j$ binary variables $v_j^l$ for
$l = 1, \ldots, k_j$, where $k_j$ is the number of distinct values this
categorical feature can take, to the MILP.
Furthermore, we add the constraint $\sum_l v_j^l = 1$ for every categorical
variable, thereby ensuring that it takes on one and only one of its values.

With these variables, each term can be directly encoded as it consists of a
linear function.
Likewise, the comparison constraints $=$, $\leq$, and $<$ can be directly
encoded in the MILP\@.
We encode $t \neq t'$ as
$\left( t < t' \right) \lor \left( t' < t \right)$ for continuous variables and
as
$\bigvee_{l \neq t'} t = l$ for categorical variables.

Next, we consider the case where $\phi$ is a boolean combination of constraints
$\phi' \land \phi''$ or $\phi' \lor \phi''$.
The first case can be encoded straightforwardly in the MILP\@.
To encode the disjunction $\phi' \lor \phi''$ we add two additional binary
variables $v'$ and $v''$ to the MILP with the constraints
\begin{align*}
    v' = 1 &\iff \phi',\\
    v'' = 1 &\iff \phi'',\\
    v' + v'' &\geq 1.
\end{align*}

Finally, if $\phi$ is a negation $\lnot \phi'$ of $\phi'$, the constraint is
preprocessed and rewritten into a logically equivalent constraint before
encoding as MILP:
\begin{align*}
    \lnot \left( t = t' \right) &:= t \neq t',\\
    \lnot \left( t \leq t' \right) &:= t' < t,\\
    \lnot \left( t \neq t' \right) &:= t = t',\\
    \lnot \left( t < t' \right) &:= t' \leq t,\\
    \lnot \left( \phi' \land \phi'' \right) &:= \lnot \phi' \lor \lnot \phi'',\\
    \lnot \left( \phi' \lor \phi'' \right) &:= \lnot \phi' \land \lnot \phi'',\\
    \lnot \left( \lnot \phi' \right) &:= \phi'.
\end{align*}

\section{Individual Fairness Certificate}
\label{sec:fairness-certificate}

In this section, we prove the correctness of our individual fairness certificate
as formalized in~\cref{thm:certificate}, which allows the data consumer to prove
individual fairness of the end-to-end model $M$, once given the latent
representation $z$ and radius $\epsilon$ by the data producer:

\begin{theorem}{(Individual fairness certificate)} \label{thm:certificate}
    Suppose $M = h_\psi \circ f_\theta$ with data point $x$ and similarity
    notion $\phi$. Furthermore, let $z = f_\theta(x)$,
    $S_\phi(x) = \{x' \in \mathbb{R}^n \mid \phi(x, x')\}$ and
    $\epsilon = \max_{x' \in S_\phi(x)} ||z - f_\theta(x')||_\infty$.
    If
    \begin{equation*}
        \max_{z' \in \mathbb{B}_\infty(z, \epsilon)}
        h^{( y' )}_\psi(z') - h^{\left( y \right)}_\psi(z') < 0
    \end{equation*}
    for all labels $y'$ different from the true label $y$, then for all
    $x' \in S_\phi(x)$ we have $M(x) = M(x')$.
\end{theorem}

\begin{proof}
    The data producer computes the latent representation $z = f_\theta(x)$ and
    certifies that
    \begin{equation}
        \label{eq:epsilon}
        \epsilon = \max_{x' \in S_\phi(x)} ||z - f_\theta(x')||_\infty.
    \end{equation}
    Thus, it immediately follows that
    $f_\theta \left( S_\phi \left( x \right) \right) \subseteq
    \mathbb{B}_\infty \left( z, \epsilon \right)$, where
    $\mathbb{B}_\infty \left( z, \epsilon \right)$ is the
    $\ell_\infty$-bounding box with center $z$ and radius $\epsilon$.
    Consider any label $y'$ different from the true label $y = M(x)$.
    If the data consumer certifies that
    \begin{equation}
        \label{eq:fairness}
        \max_{z' \in \mathbb{B}_\infty(z, \epsilon)}
        h^{( y' )}_\psi(z') - h^{\left( y \right)}_\psi(z') < 0,
    \end{equation}
    then the classifier will predict label $y$ for all
    $z' \in \mathbb{B}_\infty(z, \epsilon)$.
    Combining this with
    $f_\theta \left( S_\phi \left( x \right) \right) \subseteq
    \mathbb{B}_\infty \left( z, \epsilon \right)$
    we have
    \begin{equation*}
        \forall x' \in S_\phi(x) : M(x) = M(x'),
    \end{equation*}
    implying that the end-to-end classifier is individually fair for similarity
    notion $\phi$ at data point $x$.
    We refer to~\cref{sec:certifying} and \citet{tjeng2017evaluating}
    for details on the correctness of the certificates
    for~\cref{eq:epsilon,eq:fairness}.
\end{proof}

\section{Datasets}
\label{sec:datasets}

In this section, we provide a detailed overview of the datasets considered
in~\cref{sec:experiments}.
We recall that we perform the following preprocessing on all datasets: (i)
normalize numerical attributes to zero mean and unit variance, (ii) one-hot
encode categorical features, (iii) drop rows and columns with missing values,
and (iv) split into train, test and validation sets.
Although we only consider datasets with binary classification tasks, we note
that our method straightforwardly extends to the multiclass case.

\paragraph{Adult}

The Adult Income dataset~\cite{dua2019uci} is extracted from the 1994 US Census
database.
Every sample represents an individual and the goal is to predict whether that
person's income is over 50K\$ / year.

\paragraph{Compas}

The COMPAS Recidivism Risk Score dataset contains data collected on the use of
the COMPAS risk assessment tool in Broward County, Florida
Angwin~\cite{angwin2016machine}.
The task is to predict recidivism within two years for all individuals.

\paragraph{Crime}

The Communities and Crime dataset~\cite{dua2019uci} contains socio-economic,
law-enforcement, and crime data for communities within the US\@.
We try to predict whether a specific community is above or below the median
number of violent crimes per population.

\paragraph{German}

The German Credit dataset~\cite{dua2019uci} contains 1000 instances describing
individuals who are either classified as good or bad credit risks.

\paragraph{Health}

The Heritage Health dataset (\url{https://www.kaggle.com/c/hhp}) contains
physician records and insurance claims.
For every patient we try to predict ten-year mortality by binarizing the
Charlson Index, taking the median value as a cutoff.

\paragraph{Law School}

This dataset from the Law School Admission Council's National Longitudinal Bar
Passage Study~\cite{wightman1998lsac} has application records for 25 different
law schools.
The task is to predict whether a student passes the bar exam.

We note that for some of these datasets the label distribution is highly
unbalanced as displayed in~\cref{tab:datasets}.
For example, for the Law School dataset, learning a representation that maps all
individuals to the same point in the latent space and classifying that point as
negative would yield 73.7\% test set accuracy.
Moreover, individual fairness would be trivially satisfied for any constraint
$\phi$ as all individuals are mapped to the same outcome.
It is thus important to compare the performance of all models with the base
rates from~\cref{tab:datasets}.
Moreover, for every table containing accuracy values we provide an analogous
table with balanced accuracy in~\cref{sec:balanced-accuracy}.

\begin{table}
    \caption{
    Statistics for train, validation, and test datasets.
    Note that most of the datasets, namely Adult, German, Health, and Law
    School, have a highly skewed distribution of positive labels.
    }
    \label{tab:datasets}
    \begin{center}
        \begin{small}
            \begin{sc}
                \resizebox{0.8\columnwidth}{!}{ \begin{tabular}{ccccccc}
    \toprule
    & \multicolumn{2}{c}{Train} & \multicolumn{2}{c}{Validation} & \multicolumn{2}{c}{Test} \\
    & Size & Positive & Size & Positive & Size & Positive \\
    \midrule
    Adult & 24129 & 24.9\% & 6033 & 24.9\% & 15060 & 24.6\% \\
    Compas & 3377 & 52.3\% & 845 & 52.2\% & 1056 & 55.6\% \\
    Crime & 1276 & 48.7\% & 319 & 55.5\% & 399 & 49.6\% \\
    German & 640 & 70.5\% & 160 & 66.9\% & 200 & 71.0\% \\
    Health & 139785 & 68.0\% & 34947 & 68.6\% & 43683 & 68.0\% \\
    Law School & 5053 & 27.3\% & 13764 & 26.8\% & 17205 & 26.3\% \\
    \bottomrule
\end{tabular}
 }
            \end{sc}
        \end{small}
    \end{center}
\end{table}

\paragraph{Fair Transfer Learning}

We follow \citet{madras2018learning} and consider the Health dataset for
transferable representation learning.
The original task for the Health dataset is to predict the Charlson Index.
Thus, to demonstrate transferability, we omit the primary condition group labels
from the set of features, and try to predict them from the latent representation
without explicitly optimizing for the task.
We display the (highly imbalanced) label distributions for the considered
primary condition groups in~\cref{tab:transfer-base-rates}.

\begin{table}[t]
    \caption{
        Percentage of positive labels for train, validation, and test datasets
        for transfer learning tasks.
        Note, that the percentages do not sum to 100\% as the labels are
        aggregated by patient and year.
    }
    \label{tab:transfer-base-rates}
    \begin{center}
        \begin{small}
            \begin{sc}
                \resizebox{0.5\columnwidth}{!}{ \begin{tabular}{cccc}
    \toprule
    & \multicolumn{3}{c}{Positive (\%)} \\
    & Train & Validation & Test \\
    \midrule
    msc2a3 & 62.0 & 61.9 & 61.9 \\
    metab3 & 34.9 & 34.9 & 34.9 \\
    arthspin & 31.5 & 31.7 & 32.1 \\
    neument & 28.4 & 28.5 & 28.6 \\
    respr4 & 27.5 & 27.5 & 27.5 \\
    \bottomrule
\end{tabular}
 }
            \end{sc}
        \end{small}
    \end{center}
\end{table}

\section{Experiment Setup}
\label{sec:experiment-setup}

Here, we provide a detailed overview of the model architectures and
training hyperparameters considered in~\cref{sec:experiments}.
Recall that we model the encoder $f_\theta$ as a neural network, and we use
logistic regression as a classifier $h_\psi$.
We run a grid search over model architectures and loss balancing factors
$\gamma$ which we evaluate on the validation set.
Concretely, we search over two different encoders (both with latent space of
dimension 20): (i) without a hidden layer and (ii) with a single hidden layer
of 20 neurons, and loss balancing factors $\gamma \in [10, 1, 0, 0.01]$.
As a result, we consider $f_\theta$ with one hidden layer of 20 neurons (except
for Law School where we do not have a hidden layer) and a latent space of
dimension 20.
We fix $\gamma$ to 10 for Adult, Crime, and German, to 1 for Compas and Health,
and to 0.1 for Law School.
We train our models for 100 epochs with a batch size of 256.
We use the Adam optimizer~\cite{kingma2015adam} with weight decay 0.01 and
dynamic learning rate scheduling based on validation measurements
(ReduceLROnPlateau from~\cite{paszke2019pytorch}) starting at 0.01 with a
patience 5 of epochs.
Finally, we run DL2 with 25 PGD iterations with step size 0.05 to find
counterexamples (\cf~\cref{sec:training}).

\section{Constraints}
\label{sec:constraints}

In this section, we provide a full formalization of the similarity constraints
considered in~\cref{sec:experiments}.

\paragraph{Noise (\textsc{Noise})}

Under this constraint, two individuals are similar if their normalized numerical
features differ by no more than $\alpha$.
We consider $\alpha = 0.3$ for all experiments, which means \eg for Adult: two
individuals are similar if their age difference is smaller than roughly 3.95
years.

\paragraph{Categorical (\textsc{Cat})}

We consider two individuals similar if they are identical except for one or
multiple categorical attributes.
For Adult and German, we choose the binary attribute gender.
For Compas, two people are to be treated similarly regardless of race.
For Crime, we enforce the constraint that the state should not affect prediction
outcome for two neighborhoods.
For Health, two identical patients, except for gender and age, should observe
the same ten-year mortality at their first insurance claim.
For Law School, we consider two individuals similar regardless of their race and
gender.

\paragraph{Categorical and noise (\textsc{Cat + Noise})}

This constraint combines the two previous constraints and considers two
individuals as similar if their numerical features differ no more than $\alpha$
regardless of their values for certain categorical attributes.

\paragraph{Conditional attributes (\textsc{Attribute})}

In this case, $\phi$ is composed of a disjunction of two mutually exclusive
cases, one of which has to hold for similarity.
For this, we consider a numerical attribute and a threshold $\tau$.
If two individuals are both below $\tau$, then they are similar if their
normalized attribute differences are less than $\alpha_1$.
If both individuals are above $\tau$, similarity holds if the attribute
differences are less than $\alpha_2$.
Concretely, consider two applicants from the Law School dataset.
If both of their GPAs are below $\tau = 3.4$ (the median), then they are similar
only if their difference in GPA is less than 0.1694 ($\alpha_1 = 0.4$).
However, if both their GPAs are above $3.4$, then we consider the applicants
similar if their GPAs differ less than 0.847 ($\alpha_2 = 0.2$).
For Adult, we consider the median age as threshold $\tau = 37$, with $\alpha_1 =
0.2$ and $\alpha_2 = 0.4$ which corresponds to age differences of 2.63 and 5.26
years respectively.
For German, we also consider the median age as threshold $\tau = 33$, with
$\alpha_1 = 0.2$ and $\alpha_2 = 0.4$ which corresponds to age differences of
roughly 0.24 and 0.47 years respectively.

\paragraph{Subordination (\textsc{Quantiles})}

We follow \citet{lahoti2019operationalizing} and define a constraint that
counters subordination between social groups.
We consider the Law School dataset and differentiate two social groups by race,
one group containing individuals of white race and the other containing all
remaining races.
To counter subordination, we compute within-group ranks based on the GPAs and
define similarity if the rank difference for two students from different groups
is less than 24.
Thus, two students are considered similar if their performance relative to their
group is similar even though their GPAs may differ significantly.

\section{Scaling to Large Networks}
\label{sec:scaling}

To show that our method can be easily scaled to larger networks, we train an
encoder $f_\theta$ with 200 hidden neurons and latent space dimension 200.
For such large models we can relax the MILP encodings to a linear
program~\cite{ehlers2017formal} and solve for robustness via convex relaxation.
Running this relaxation for our large network and the \textsc{Noise} constraint
on Adult we can certify fairness for 91.4\% of the individuals with
82.8\% accuracy and average certification runtime of 1.13s.
In contrast, the complete solver can certify 92.6\% of individuals with average
runtime of 31.9s.
For even larger model architectures, one can use one of the recent
state-of-the-art network verifiers~\cite{singh2019krelu}.

\section{Balanced Accuracy}
\label{sec:balanced-accuracy}

We recall that some of the datasets are highly imbalanced (\cf
\cref{tab:datasets}).
Hence, we evaluate the balanced accuracies for the models
from~\cref{tab:constraints} and show them in~\cref{tab:constraints-balanced}.
It can be observed that LCIFR performs only slightly worse than the baseline
across all constraints and datasets (except for \textsc{Cat + Noise} on German).

\begin{table}
    \caption{
        Balanced accuracy for encoders and classifiers
        from~\cref{tab:constraints}.
    }
    \label{tab:constraints-balanced}
    \begin{center}
        \begin{small}
            \begin{sc}
                \resizebox{0.65\columnwidth}{!}{ \begin{tabular}{cccc}
    \toprule
    & & \multicolumn{2}{c}{Balanced Accuracy (\%)} \\
    Constraint & Dataset & Base & LCIFR \\
    \midrule
    \multirow{6}{*}{Noise} & Adult & 74.5 & 70.9 \\
    & Compas & 65.1 & 62.3 \\
    & Crime & 84.4 & 83.2 \\
    & German & 69.6 & 60.8 \\
    & Health & 77.1 & 76.5 \\
    & Law School & 76.1 & 75.8 \\
    \midrule
    \multirow{6}{*}{Cat} & Adult & 74.7 & 73.9 \\
    & Compas & 64.9 & 65.7 \\
    & Crime & 84.4 & 83.9 \\
    & German & 69.2 & 68.3 \\
    & Health & 77.2 & 77.1 \\
    & Law School & 76.1 & 75.5 \\
    \midrule
    \multirow{6}{*}{Cat + Noise} & Adult & 74.7 & 70.8 \\
    & Compas & 64.9 & 62.5 \\
    & Crime & 84.4 & 81.7 \\
    & German & 69.2 & 49.8 \\
    & Health & 77.2 & 76.5 \\
    & Law School & 76.1 & 75.5 \\
    \midrule
    \multirow{3}{*}{Attribute} & Adult & 74.5 & 70.1 \\
    & German & 69.6 & 61.8 \\
    & Law School & 76.1 & 74.3 \\
    \midrule
    Quantiles & Law School & 76.1 & 75.8 \\
    \bottomrule
\end{tabular}
 }
            \end{sc}
        \end{small}
    \end{center}
\end{table}

\paragraph{Fair Transfer Learning}

We recall that the label distribution of the primary condition groups (transfer
tasks) are highly imbalanced (\cf \cref{tab:transfer-base-rates}).
Nevertheless, LCIFR achieves accuracies that are above the base rate achieved by
majority class prediction (\cf~\cref{tab:constraints}) in all cases except for
\textsc{RESPR4}.
Here, we display the corresponding balanced accuracies
in~\cref{tab:transfer-balanced}, and we observe that the balanced accuracies are
inversely proportional to the label imbalance
(\cf~\cref{tab:transfer-base-rates}).

\begin{table}
    \caption{
        Balanced accuracy for transferable representation learning on Health
        dataset with \textsc{Cat + Noise} constraint from~\cref{tab:transfer}.
    }
    \label{tab:transfer-balanced}
    \begin{center}
        \begin{small}
            \begin{sc}
                \resizebox{0.65\columnwidth}{!}{ \begin{tabular}{ccc}
    \toprule
    Task & Label & Balanced Accuracy (\%) \\
    \midrule
    Original & Charlson Index & 63.9 \\
    \midrule
    \multirow{5}{*}{Transfer} & msc2a3 & 70.8 \\
    & metab3 & 68.5 \\
    & arthspin & 66.0 \\
    & neument & 58.9 \\
    & respr4 & 56.0 \\
    \bottomrule
\end{tabular}
 }
            \end{sc}
        \end{small}
    \end{center}
\end{table}

    }{}

    \message{^^JLASTPAGE \thepage^^J}

\end{document}